\documentclass[lettersize,journal]{IEEEtran}
\usepackage{amsmath,amsfonts}
\usepackage{mathtools}
\DeclareMathOperator*{\argmax}{arg\,max}
\usepackage{algorithmic}
\usepackage{algorithm}
\usepackage{array}
\usepackage{cleveref}
\usepackage[caption=false,font=footnotesize]{subfig}
\usepackage{pdfpages}

\usepackage{textcomp}
\usepackage{stfloats}
\usepackage{url}
\usepackage{verbatim}
\usepackage{graphicx}

\usepackage{booktabs}
\usepackage{cite}
\usepackage{tikz}
\usetikzlibrary{arrows.meta,positioning,calc,fit,backgrounds,shadows.blur}
\usepackage{xcolor}

\definecolor{pastelBlue}{HTML}{BFD7EA}
\definecolor{pastelGreen}{HTML}{CDECCF}
\definecolor{pastelOrange}{HTML}{F9DCC4}
\definecolor{pastelPurple}{HTML}{D5C4E1}
\definecolor{pastelGray}{HTML}{EAEAEA}

\tikzset{
  >={Latex[length=2.5mm]},
  block/.style={draw=black, very thick, rounded corners=3pt,
                align=center, minimum height=12mm, minimum width=42mm,
                blur shadow},
  smallblock/.style={block, minimum width=38mm, minimum height=10mm},
  io/.style={block, fill=pastelBlue},
  proc/.style={block, fill=pastelGreen},
  core/.style={block, fill=pastelOrange},
  outnode/.style={block, fill=pastelPurple, minimum width=40mm},
  note/.style={draw=black!60, fill=pastelGray, rounded corners=3pt,
               inner sep=3pt, align=left, font=\footnotesize},
  link/.style={ultra thick, draw=black!80},
  thinlink/.style={thick, draw=black!70, dashed},
}

\usepackage{tikz}
\usepackage{graphicx}
\usepackage{mwe} % for example-image-a/b/c/d placeholders
\usetikzlibrary{matrix,positioning,fit,arrows.meta,calc,shadows.blur}
\definecolor{pastelBlue}{HTML}{BFD7EA}
\definecolor{pastelGreen}{HTML}{CDECCF}
\definecolor{pastelOrange}{HTML}{F9DCC4}
\definecolor{pastelPurple}{HTML}{D5C4E1}
\definecolor{pastelGray}{HTML}{EAEAEA}

\newcommand{\affilfoot}[2]{%
  \noindent\hangindent=1.8em\hangafter=1%
  \hbox to 1.5em{$^{#1}$\hfil}#2\par
}

\begin{document}

\title{Modeling Wise Decision Making: A Z-Number Fuzzy \\ Framework Inspired by Phronesis}

\author{Sweta Kaman$^{1}$, Ankita Sharma$^{2}$, and Romi Banerjee$^{3}$%
\thanks{%
\noindent\rule{\columnwidth}{0.4pt}\par
\textbf{Corresponding author:} Sweta Kaman (kaman.1@iitj.ac.in)\par
\medskip
\textbf{Sweta Kaman}\par
kaman.1@iitj.ac.in\par
\textbf{Ankita Sharma}\par
ankitasharma@iitj.ac.in\par
\textbf{Romi Banerjee}\par
romibanerjee@iitj.ac.in\par
\medskip
\affilfoot{1}{School of AI and Data Science, Indian Institute of Technology Jodhpur, \textbf{Jodhpur}, India}%
\affilfoot{2}{School of Liberal Arts, Indian Institute of Technology Jodhpur, \textbf{Jodhpur}, India}%
\affilfoot{3}{Department of Computer Science and Engineering, Indian Institute of Technology Jodhpur, \textbf{Jodhpur}, India}%
}%
}

\maketitle

\begin{abstract}
\textbf{Background:} Wisdom is a superordinate construct that embraces perspective taking, reflectiveness, prosocial orientation, reflective empathetic action, and intellectual humility. Unlike conventional models of reasoning that are rigidly bound by binary thinking, wisdom unfolds in shades of ambiguity, requiring both graded evaluation and self-reflective humility. Current measures depend on self-reports and seldom reflect the humility and uncertainty inherent in wise reasoning. A computational framework that takes into account both multidimensionality and confidence has the potential to improve psychological science and allow humane AI.   

\textbf{Method:} We present a fuzzy inference system with Z numbers, each of the decisions being expressed in terms of a wisdom score (restriction) and confidence score (certainty). As part of this study, participants ($N=100$) were exposed to culturally neutral pictorial moral dilemma tasks to which they generated think-aloud linguistic responses, which were mapped into five theoretically based components of wisdom. The scores of each individual component were combined using a base of 21 rules, with membership functions tuned via Gaussian kernel density estimation. 

\textbf{Results:} In a proof of concept study, the system produced dual attribute wisdom representations that correlated modestly but significantly with established scales ($\rho \approx .20$–.22) while showing negligible relations with unrelated traits, supporting convergent and divergent validity. 

\textbf{Contribution:} The contribution is to formalize wisdom as a multidimensional, uncertainty-conscious construct, operationalized in the form of Z-numbers. In addition to progressing measurement in psychology, it calculates how fuzzy Z numbers can provide AI systems with interpretable, confidence-sensitive reasoning that affords a safe, middle ground between rigorous computation and human-like judgment.  
\end{abstract}


\section{Introduction}

What makes a \emph{wise} decision different from a \emph{correct} one? It is not just the result that is important, but the complex process of striking a balance between opposing values, adopting the views of others, addressing ambiguity, and translating confidence in belief into intellectual humility. Think of the daily actions challenging this trait: the decision to take a found wallet back, when you feel you are too busy, or when a child is subjected to cruelty; or a move to stop the waste of valuable water. What defines wisdom in such cases is not only the moral intention but also the ability to weigh these factors and arrive at a decision appropriate for the specific situation, integrating judgment with context.

Our methodology is deeply aligned with the definition of phronesis or practical wisdom, as originally expressed by Aristotle in the Nicomachean Ethics \cite{aristotle350}, where he contrasted practical wisdom with theoretical knowledge (episteme) and technical skill (techne). Phronesis deals with the ability to think wisely in regards to what should be done, along the way of moral uncertainty, and contextual complexity to achieve human flourishing.  In this regard, our Z-numbers framework resembles the Aristotle vision: Attribute A embodies the structure of wise reasoning on a multitude of attributes whereas Attribute B is the self-knowledge and the feeling of surety in oneself, which taken together provides an analog of phronesis.

Wisdom, investigated in philosophy and psychology, is a fabric of cognitive, reflective, affective, and sometimes moral components
\cite{baltes2000, ardelt2003, sternberg1998, sternberg1990wisdom,staudinger2011psychological,ardelt2006,glueck2019}. Human wisdom, which is difficult to measure, has been assessed using various methodological approaches. Performance-based paradigms, such as the Berlin Wisdom Paradigm \cite{bwp} and Bremen Wisdom Paradigm \cite{brwp}, evaluate the reasoning of participants in structured tasks. Recently, hybrid instruments such as the Situated Wise Reasoning Scale (SWIS) \cite{brienza2018situated} have combined scenario-based reflection with self-report items to bridge the gap between performance-based and questionnaire methods. In addition, several self-report scales have been developed, including (but not limited to) the Three-Dimensional Wisdom Scale (3D-WS) \cite{ardelt2003}, the Self-Assessed Wisdom Scale (SAWS) \cite{webster2003}, and the San Diego Wisdom Scale (SD-WISE) \cite{jeste2017}. While these instruments have advanced the field, self-reports remain vulnerable to social desirability, biased self-perception, and a limited ability to capture the behavioral expressions of wisdom. 

Importantly, wisdom cannot be constricted to only one ability or component \cite{bangen2013defining}. Both philosophical and psychological literature indicate that no single constituent, whether a perspective-taking, reflectiveness or humility, can exhaustively refer to or represent wisdom \cite{zhang2023wisdom}. Rather, wisdom is a synthesis of several interrelated components, thus an inherently multi-compositional concept \cite{zhang2023wisdom}.
Building on these traditions, we conceptualized wisdom as a \textbf{superordinate construct} that manifests through multiple component processes rather than as a unitary trait. Specifically, our framework operationalizes five components that recur across integrative theories of wisdom \cite{sternberg1990wisdom,baltes2000, ardelt2003, glueck2019, brienza2018, grossmann2020}: perspective taking (PT), reflectiveness (REF), prosocial orientation (PO), reflective empathetic action (REA), and intellectual humility (IH). 

The quality of wise judgment is central to human well-being, and its effects extend to clinical decision-making, policy decision-making, dispute resolution, and the design of ethical artificial intelligence \cite{sternberg1990wisdom,staudinger2011psychological}. This framework illuminates how the precision of computational tools combined with the richness of human insight can provide a transparent, interpretable, and context-sensitive measure of wisdom. Simultaneously, \textbf{computational intelligence} has evolved to the level where it can demonstrate real-world outcomes in the form of complex, qualitative judgments made by human beings \cite{mendel2017,abraham2005}.

Fuzzy logic offers a more humane and cognitively realistic formalism, because it allows degrees of truth to be represented and not absolute categories. This has been important in modeling decisions where the judgment is very much in the grey, one where situational sensitivity and subjective interpretation play a role, which is what is involved in the psychological as well as philosophical traditions of wisdom.

The Z-number introduced by Zadeh \cite{zadeh2011} expands upon fuzzy logic by assigning to every value a measure of confidence, which is very useful when representing uncertainty. This Z-number framework is meaningfully operationalized in this work within a Mamdani fuzzy inference system \cite{mamdani1974} where the rules describing the outcomes of moral reasoning made by participants are interpretable, but also retain the complexity and humility of human wisdom. In contrast to probabilistic or type-2 fuzzy systems, which deal with uncertainty but tend to lose semantics, the Z-number system considers both valuation and confidence within one and the same formalism. They are especially appropriate to the modeling of wisdom, at which judgments can be about not only the content that is decided, but also the strength of that judgment.

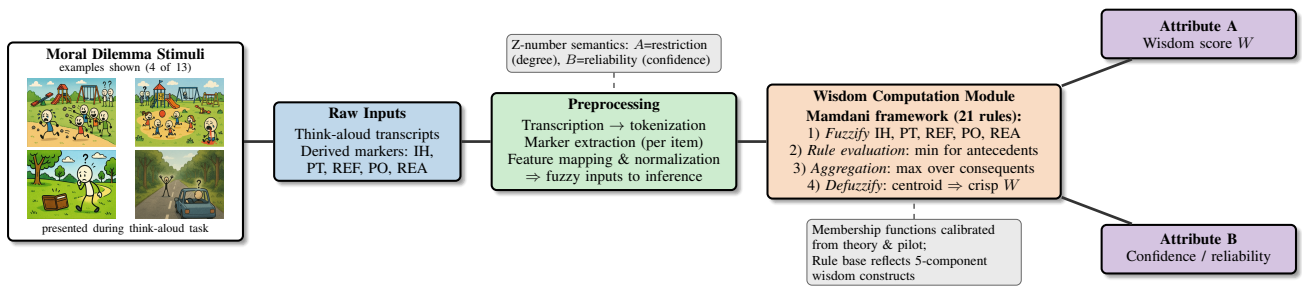
\begin{figure*}[!t]
\centering
\resizebox{.95\textwidth}{!}{%
\begin{tikzpicture}[node distance=10mm and 6mm, font=\small]
\tikzset{
  >={Latex[length=2.5mm]},
  block/.style={draw=black, very thick, rounded corners=3pt,
                align=center, minimum height=10mm, minimum width=34mm,
                blur shadow},
  smallblock/.style={block, minimum width=30mm, minimum height=9mm},
  io/.style={block, fill=pastelBlue, text width=36mm},
  proc/.style={block, fill=pastelGreen, text width=48mm},
  core/.style={block, fill=pastelOrange, text width=48mm},
  outnode/.style={block, fill=pastelPurple, minimum width=40mm}, % <-- renamed
  note/.style={draw=black!60, fill=pastelGray, rounded corners=3pt,
               inner sep=3pt, align=left, font=\footnotesize, text width=42mm},
  link/.style={ultra thick, draw=black!80},
  thinlink/.style={thick, draw=black!70, dashed},
}

\node[block, fill=white, text width=46mm, minimum height=32mm] (stim) {%
\textbf{Moral Dilemma Stimuli}\\[-1pt]
\scriptsize examples shown (4 of 13) \\[3pt]
\begin{tabular}{@{}cc@{}}
\includegraphics[width=18mm,height=14mm]{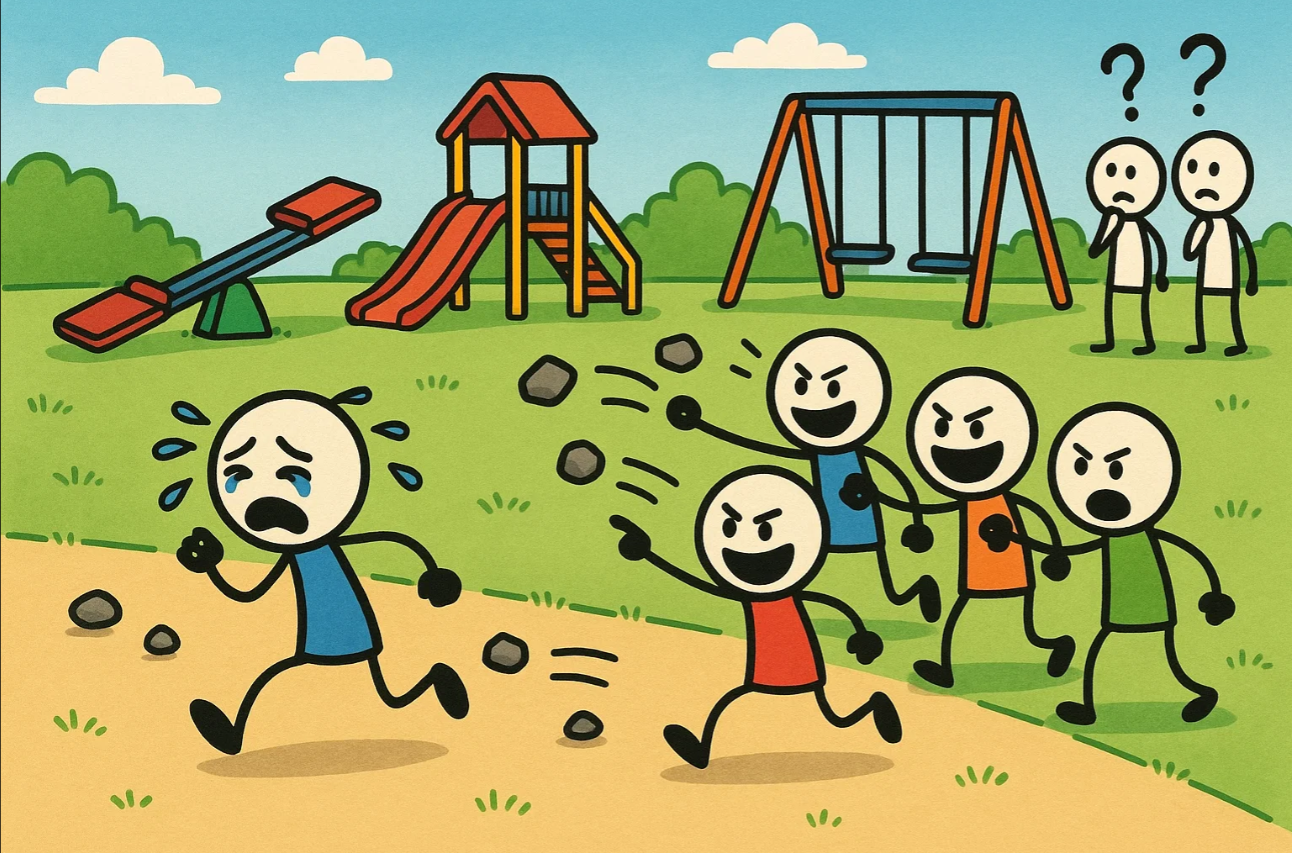} &
\includegraphics[width=18mm,height=14mm]{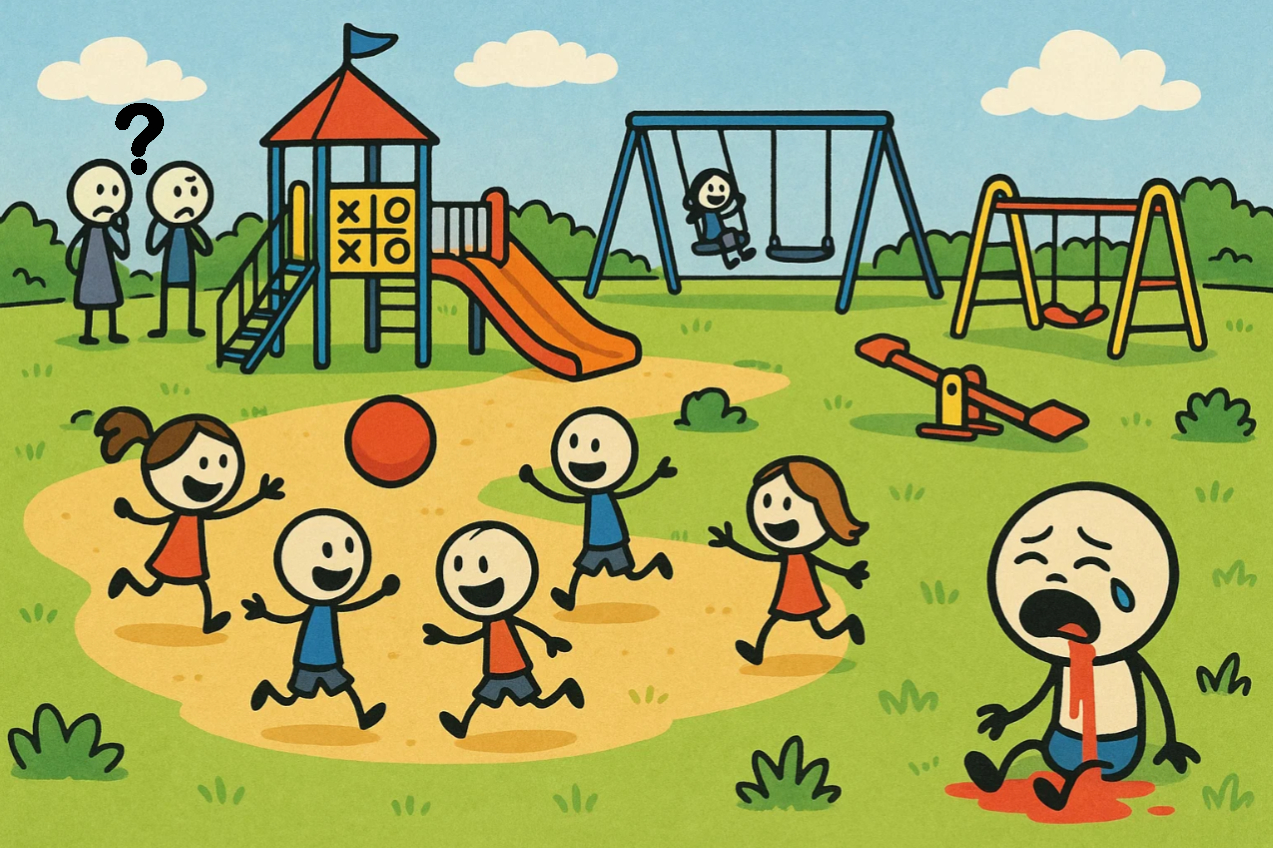} \\
\includegraphics[width=18mm,height=14mm]{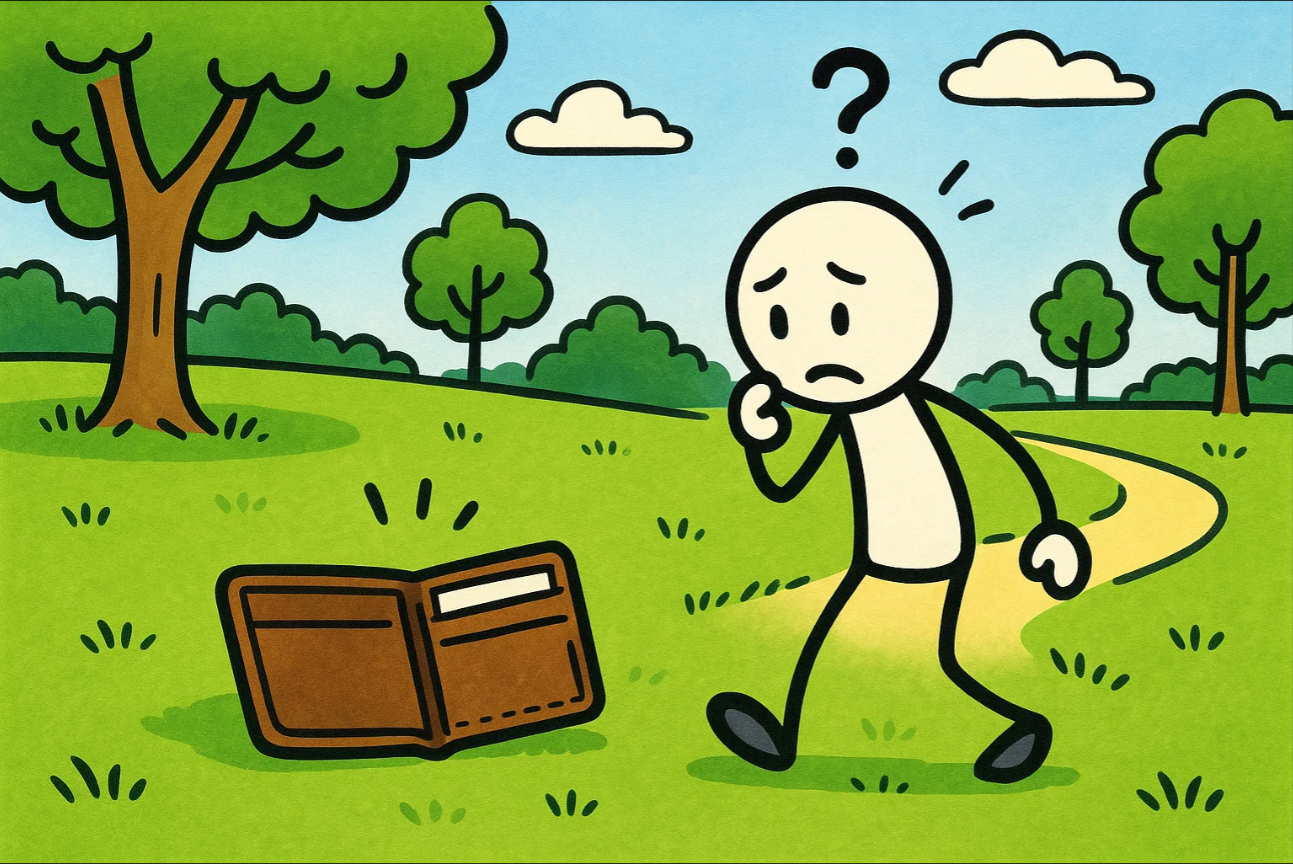} &
\includegraphics[width=18mm,height=14mm]{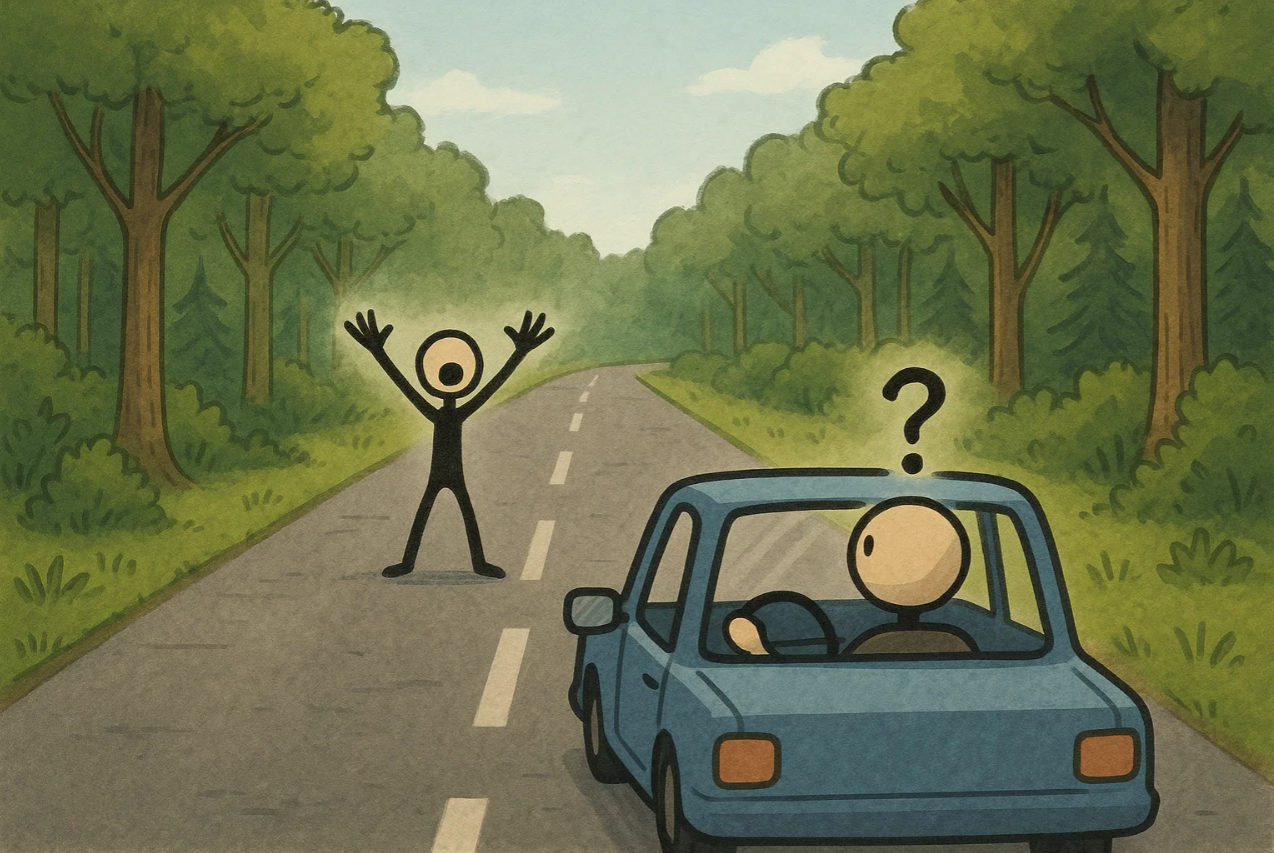} \\
\end{tabular}\\[1pt]
\scriptsize presented during think-aloud task
};

\node[io, right=of stim] (inputs) {%
\textbf{Raw Inputs}\\[2pt]
Think-aloud transcripts\\
Derived markers: IH, PT, REF, PO, REA
};

\node[proc, right=of inputs] (pre) {%
\textbf{Preprocessing}\\[2pt]
Transcription $\rightarrow$ tokenization\\
Marker extraction (per item)\\
Feature mapping \& normalization\\
\(\Rightarrow\) fuzzy inputs to inference
};

\node[core, right=of pre, text width=58mm] (engine) {%
\textbf{Wisdom Computation Module}\\[2pt]
\textbf{Mamdani framework (21 rules):}\\
1) \emph{Fuzzify} IH, PT, REF, PO, REA\\
2) \emph{Rule evaluation}: min for antecedents\\
3) \emph{Aggregation}: max over consequents\\
4) \emph{Defuzzify}: centroid $\Rightarrow$ crisp $W$
};

\node[outnode, above right=5mm and 8mm of engine] (Aout) {%
\textbf{Attribute A}\\
Wisdom score \(W\)
};
\node[outnode, below right=5mm and 8mm of engine] (Bout) {%
\textbf{Attribute B}\\
Confidence / reliability
};

\draw[link] (stim) -- (inputs);
\draw[link] (inputs) -- (pre);
\draw[link] (pre) -- (engine);
\draw[link] (engine) -- (Aout);
\draw[link] (engine) -- (Bout);

\node[note, above=4mm of pre] (noteZ) {%
Z-number semantics: \(A\)=restriction (degree), \(B\)=reliability (confidence)
};
\draw[thinlink] (noteZ) -- (pre.north);

\node[note, below=4mm of engine] (noteRules) {%
Membership functions calibrated from theory \& pilot;\\
Rule base reflects 5-component wisdom constructs
};
\draw[thinlink] (noteRules) -- (engine.south);

\end{tikzpicture}
}% <- keep this brace
\caption{End-to-end architecture of the Z-number wisdom pipeline. Participants view dilemma stimuli and provide think-aloud responses. Transcripts are processed to extract linguistic markers (IH, PT, REF, PO, REA), mapped and normalized, and fed to a Mamdani inference system to yield the wisdom score (Attribute~A). Reliability cues form the confidence estimate (Attribute~B), giving the final Z-number semantics \(Z=(A,B)\).}
\label{fig:znumber_arch}
\end{figure*}

This study provides a performance-based model for measuring wisdom as wise decision-making based on Z-numbers and human-mediated reasoning. The participants explored culturally neutral pictorial moral dilemmas and expressed their thoughts on decision-making processes. Five literature-derived pillars of wisdom: Perspective-Taking (PT), Reflectiveness (REF), Prosocial Orientation (PO), Reflective Empathetic Action (REA), and Intellectual Humility (IH), are where their words are mapped, which feeds into a transparent 21-rule Mamdani  engine \cite{mamdani1974} \textbf{(Fig.~\ref{fig:znumber_arch})}. The outcome is a Z-number: \emph{Attribute A}, a fuzzy score of wisdom reduced down out of such components and \emph{Attribute B}, a fuzzy measure of self-reported confidence in one’s decision. Gaussian kernel density estimation (KDE) shapes membership functions to guarantee empirical rigor without impairing the theoretical foundations.
Our core contribution is computational: we develop a Z-number fuzzy system to model wisdom as an interpretable cognitive attribute. The empirical study serves to demonstrate its feasibility and preliminary construct validity rather than psychometric validation per se.

\subsection{Objectives}
Our aims are to:  
\begin{enumerate}
    \item \textbf{Design} a transparent, expert-based rule base that operationalizes wise decision-making through five components (PT, REF, PO, REA, IH).

    \item \textbf{Integrate} data-driven membership calibration via Gaussian KDE within a Z-number framework.
    \item \textbf{Examine}  the feasibility and preliminary validity of the system using performance-based moral dilemmas.
\end{enumerate}

\subsection{Significance of the study}
The proposed model provides a number of distinctive contributions:
\begin{itemize}
    \item \textbf{Dual-Attribute Representation:}  The system provides the measurement of \textit{how wise} a decision is and \textit{how confident} is the decision-maker by generating Z-numbers. This is more comprehensive than conventional unidimensional scoring in studies of wisdom.
    \item \textbf{Data-Driven yet Theory-Aligned:} Empirical distributions are used to generate membership functions through the KDE method, and the fuzzy rule base is grounded in the peer-reviewed literature, and this relationship ensures theoretical fidelity.
    \item \textbf{Validation Across Constructs:} The scores measuring the Attribute~A in the model exhibited small but statistically significant correlations with the SDWISE, PT and SAWS subscales but negligible relationships with HEXACO traits and the scores on Raven matrices problems, respectively, showing both convergent and divergent validity.
    \item \textbf{Application Potential:} The Z-number outputs (wisdom level + confidence) allow intelligent systems to communicate both judgment and self-assessed certainty in a human-interpretable way \cite{phillips2021four}. This is crucial for real-world applications—such as clinical decision support, policy simulation, and adaptive education \cite{xu2025ai, umoke2025governance}—where transparent reasoning and acknowledgment of uncertainty are necessary for trust, accountability, and ethically sensitive AI \cite{phillips2021four, lindenmeyer2025trustworthy, umoke2025governance}.
\end{itemize}

\subsection{Paper organization}
The paper is structured as follows:

\begin{itemize}
    \item \textbf{Section II}: Review of wisdom evaluation, fuzzy systems in psychology, and Z-number theory.
    \item \textbf{Section III}: Methodology, participants, moral dilemma stimuli, and linguistic feature extraction for five wisdom components.
    \item \textbf{Section IV}: Attribute A pipeline—KDE-based membership derivation and component-level aggregation (Algorithm~\ref{alg:component_level_znumber}).
    \item \textbf{Section V}: Attribute B computation—confidence normalisation, KDE-based segmentation, and fuzzy set building.
    \item \textbf{Section VI}: Mamdani fuzzy inference framework and 21-rule base for wisdom-level Z-number inference.
    \item \textbf{Section VII}: Construct validity.
    \item \textbf{Section VIII}: Results, implications, limitations, and interdisciplinary applications.
    \item \textbf{Section IX}: Conclusion and future work.
\end{itemize}

\section{Related Work}
This work is based on three strands: (i) psychological and cultural grounds of wisdom and moral reasoning, (ii) computational modelling under uncertainty (fuzzy systems, cognitive maps, Z-numbers), and (iii) performance paradigms and validation. The persistent gaps involve the use of self-reports, absent uncertainty modeling, absent theory-grounded rules, ad-hoc membership functions, nonexistent dual-attribute outputs, limited implementation of think-alouds, restricted validation, and poor cultural generalizability.

\subsection{Psychological and Cultural Foundations}
The history of moral reasoning has dipped into philosophy--utilitarianism, deontology, and the virtues (Bentham \cite{bentham1789}, Mill \cite{mill1863}, Kant \cite{kant1785}, Aristotle \cite{aristotle350}) as well as psychology, such as the stages of moral development by Kohlberg \cite{kohlberg1969}, criticisms by Gilligan \cite{gilligan1982}, intuitionism postulated by Haidt \cite{haidt2001}, and dual-process models by Kahneman \cite{greene2004,kahneman2011}. These are mostly \emph{categorical}, with little graded uncertainty encoded, even though they are highly influential. Fuzzy logic provides a natural transition because it models gradation and meta-uncertainty \cite{kosko1994,zadeh1965}.

\textit{Cultural grounding.} 
Sociocultural work emphasizes individualism-collectivism and analytic-dialectic styles, with collectivist contexts emphasizing harmony/duty, domination of moral areas varying among societies, and dialectical acceptance of inconsistencies. These warn against universal-stage theorizing and encourage culturally condensed flexible formalizations \cite{shweder1997, markus1991, triandis1995, peng1999, spencer2004, henrich2010, malle2016}. The component structure is supported by REA, PO, IH, PT, and REF, combined into a single wisdom in Indian collectivistic contexts \cite{sharma2024_hindi} \textbf{(see Table~\ref{tab:theory_attributeA})}. The presence of neutral pictorial vignettes and \emph{think-aloud} protocols \cite{ericsson1980, ericsson1993} decreases bias and notes the situated reasoning.

\subsection{Computational Modelling Under Uncertainty: Fuzzy, FCM, Z-Numbers, and Alternatives}
Formative fuzzy models use understandable, cognitive-science-informed semantics to wisdom \cite{hoppe2012}, or ethical principles mediated and encoded via transparent causal connections represented by fuzzy cognitive maps \cite{hein2022}. These improve interpretability  and similarly remain based on non-performance data as well as low validation.

\textbf{Z-numbers and membership design.} Zadeh introduced Z-numbers \cite{zadeh2011}, which assign valuation to values and confidence. Extensions involve ranking and decision-making structures \cite{banerjee2022, liao2024} applied to reliability and group choice \cite{aghaei2021, anjaria2022, chai2023}. However, they are not used in wisdom contexts as dual-attribute outputs. Similarly, membership calibration is rarely principled or data-driven (for example KDE) in psychology-facing systems, and is heuristic.

\textbf{Alternatives.} Other uncertainty frameworks, such as probabilistic reasoning \cite{pearl1988}, dempster-shafer theory \cite{shafer1976}, and interval type-2 fuzzy logic \cite{mendel2017}, manage imprecision at the cost of sacrificing either semantic transparency or psychological mapping. This also requires that the methods are mathematically rigorous but theoretically consistent.

\begin{table}[!t]
\scriptsize
\centering
\caption{Selected Foundations Informing Attribute~A (Wisdom Content)}
\label{tab:theory_attributeA}
\renewcommand{\arraystretch}{1.15}
\setlength{\tabcolsep}{3.5pt}
\begin{tabular}{p{0.32\columnwidth} p{0.50\columnwidth}}
\toprule
\textbf{Source} & \textbf{Core Idea $\rightarrow$ Relevance} \\
\midrule
Aristotle, \emph{Phronesis} \cite{aristotle350} &
Context-sensitive practical reasoning $\rightarrow$ combine cognitive and moral components. \\
Sternberg (1998) \cite{sternberg1990wisdom} &
Balancing intra/inter/extrapersonal interests $\rightarrow$ prioritise PT, PO. \\
Baltes \& Staudinger (2000) \cite{baltes2000} &
Meta-cognition, uncertainty awareness $\rightarrow$ Reflectiveness, IH. \\
Ardelt (2003) \cite{ardelt2003} &
Cognitive/reflective/affective triad $\rightarrow$ REF. \\
Gl\"uck et al. (2019) \cite{glueck2019}&
Emotion regulation, empathy $\rightarrow$ REA salience. \\
Brienza \& Grossmann (2017) \cite{brienza2017}; Grossmann et al. (2020) \cite{grossmann2020} &
Situated wise reasoning; humility central $\rightarrow$ strong weight on IH, PT. \\
\bottomrule
\end{tabular}
\end{table}

\subsection{Performance Paradigms, Think-Aloud Protocols, and Validation} Moral strategy research is informed by cognitive and decision-neuroscience models. MoralDM, which is developed by Kenneth Forbus and colleagues \cite{blass2015moral}, is a cognitive model of moral decision-making has both structured knowledge and analogical retrieval \cite{dehghani2008}, and the Moral Strategy Model relates strategy classes to neural correlates \cite{van2019}. These reinforce performance paradigms but seldom contain any explicit uncertainty, membership design based on data, dual-attribute output, or \emph{think-aloud} inputs. Moral learning reviews also report that there are other mechanisms that lack interpretable formalisms of rules or membership categories \cite{lockwood2025}.

Think-aloud protocols capture real-time reasoning and metacognition and are widely applied in moral and cross-cultural contexts \cite{guss2018,bajovic2021meta}. Best practice stresses instruction, high-quality recording, verbatim transcription, segmentation, and inter-rater reliability \cite{ericsson1993}. However, computational wisdom systems still underuse such speech data, favoring text vignettes that risk cultural bias. Validation often rests on correlations with scales or group-level tests \cite{kosinski2013}, with limited replication and construct breadth issues.   

There are still important weaknesses, including the use of self-report/vignettes, lack of confidence-aware modelling, non-theory-driven rules, heuristic thresholds, limited validation, and non-use of think-aloud protocols. We handle these through \emph{pictorial, culturally neutral dilemmas} and \emph{think-alouds} \textbf{(see \textit{Supplementary Table S1 and Figure S1})}, a \emph{Mamdani} fuzzy system \cite{mamdani1974} with 21 theory-derived rules \textbf{(see \textit{Supplementary Table S2})}, a KDE-based membership design, \emph{Z-numbers} for dual-attribute outputs. This combination of performance elicitation, data-driven semantics, theory-guided rules, and reliability-driven outputs characterizes an interpretable novel approach to computational wisdom evaluation.

\section{Methodology}

\subsection{Participants and Ethics}
We recruited $N=100$ Indian participants via purposive \cite{robinson2014} and snowball sampling \cite{goodman1961} from urban and semi-urban settings (March--December 2024). The study received institutional ethics approval (IEC/2023-24/10). Participants provided informed consent (assent with guardian consent for minors), were briefed about audio recording, confidentiality, and their right to withdraw without penalty. Although wisdom is typically studied in adulthood, we deliberately included younger participants (ages 10–17) to explore whether markers of wise reasoning emerge earlier in development. This approach avoids prematurely restricting the construct and allows a broader test of its potential age span. Core demographics are summarised in \textbf{Table~\ref{tab:participant_demographics}}. Individual sessions lasted 15--20 minutes. All procedures complied with the Declaration of Helsinki and local IRB regulations \cite{helsinki}.

\begin{table}[!t]
\scriptsize
\centering
\caption{Participant Demographics (N=100)}
\label{tab:participant_demographics}
\renewcommand{\arraystretch}{1.15}
\setlength{\tabcolsep}{5.2pt}
\begin{tabular}{lcc}
\toprule
\textbf{Characteristic} & \textbf{$n$} & \textbf{\% of $N$} \\
\midrule
\multicolumn{3}{l}{\emph{Age (years)}} \\
\quad Mean $\pm$ SD \hspace{1.5em}(Median) & \multicolumn{2}{c}{23.48 $\pm$ 5.15 \ (23.5)} \\
\quad Range & \multicolumn{2}{c}{10--39} \\
\quad 10--17 & 12 & 12.0 \\
\quad 18--20 & 17 & 17.0 \\
\quad 21--29 & 58 & 58.0 \\
\quad 30--39 & 13 & 13.0 \\
\midrule
\multicolumn{3}{l}{\emph{Gender}} \\
\quad Female & 53 & 53.0 \\
\quad Male & 47 & 47.0 \\
\midrule
\multicolumn{3}{l}{\emph{Category}} \\
\quad School Students & 18 & 18.0 \\
\quad STEM Students & 41 & 41.0 \\
\quad Non-STEM Students & 37 & 37.0 \\
\quad Working Professionals & 4 & 4.0 \\
\bottomrule
\end{tabular}
\end{table}

\subsection{Moral Decision-Making Paradigm and Pictorial Stimuli}
\begin{figure*}[t]
\centering
\subfloat[Child bullying]{%
  \includegraphics[height=0.156\textheight,keepaspectratio]{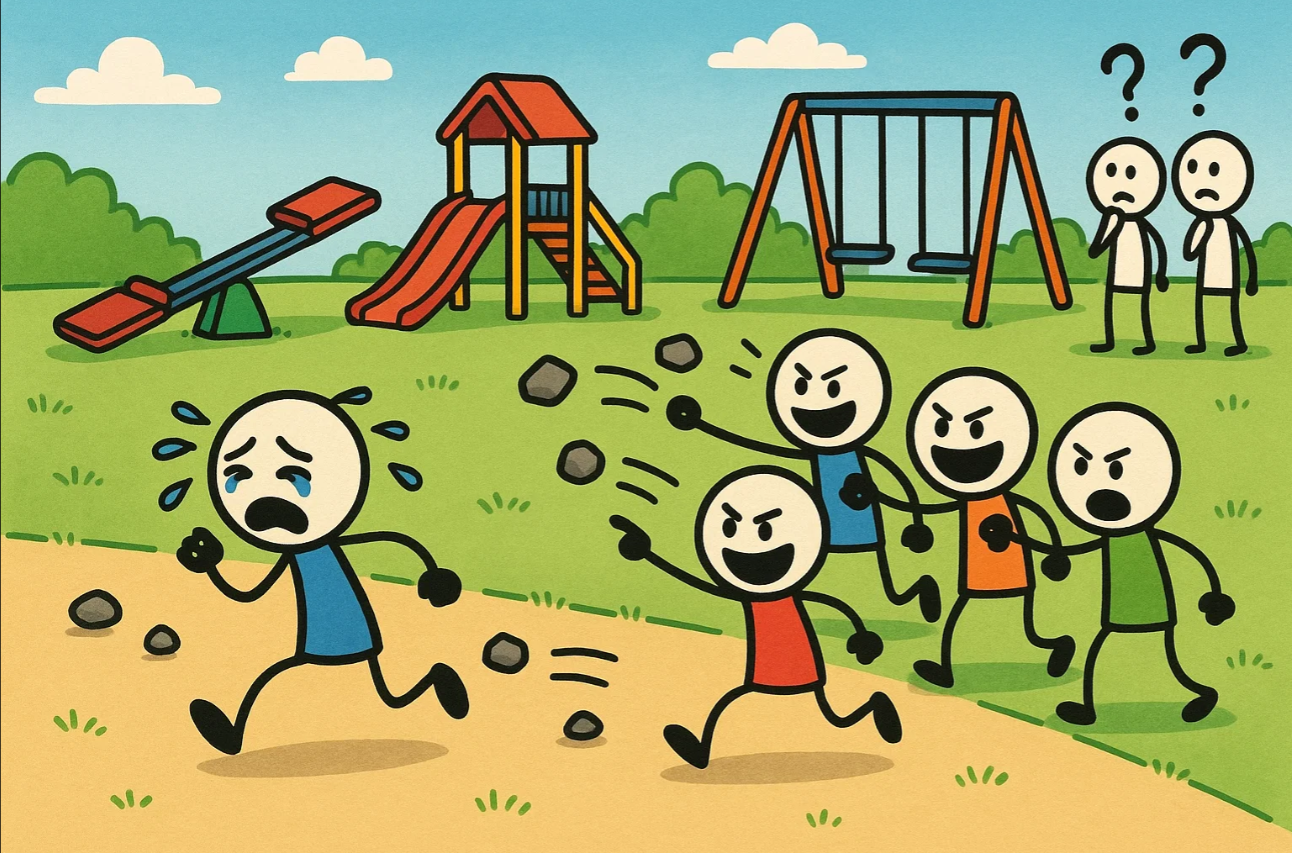}%
  \label{fig:stim1}}
\hfil
\subfloat[Helping a stranger]{%
  \includegraphics[height=0.156\textheight,keepaspectratio]{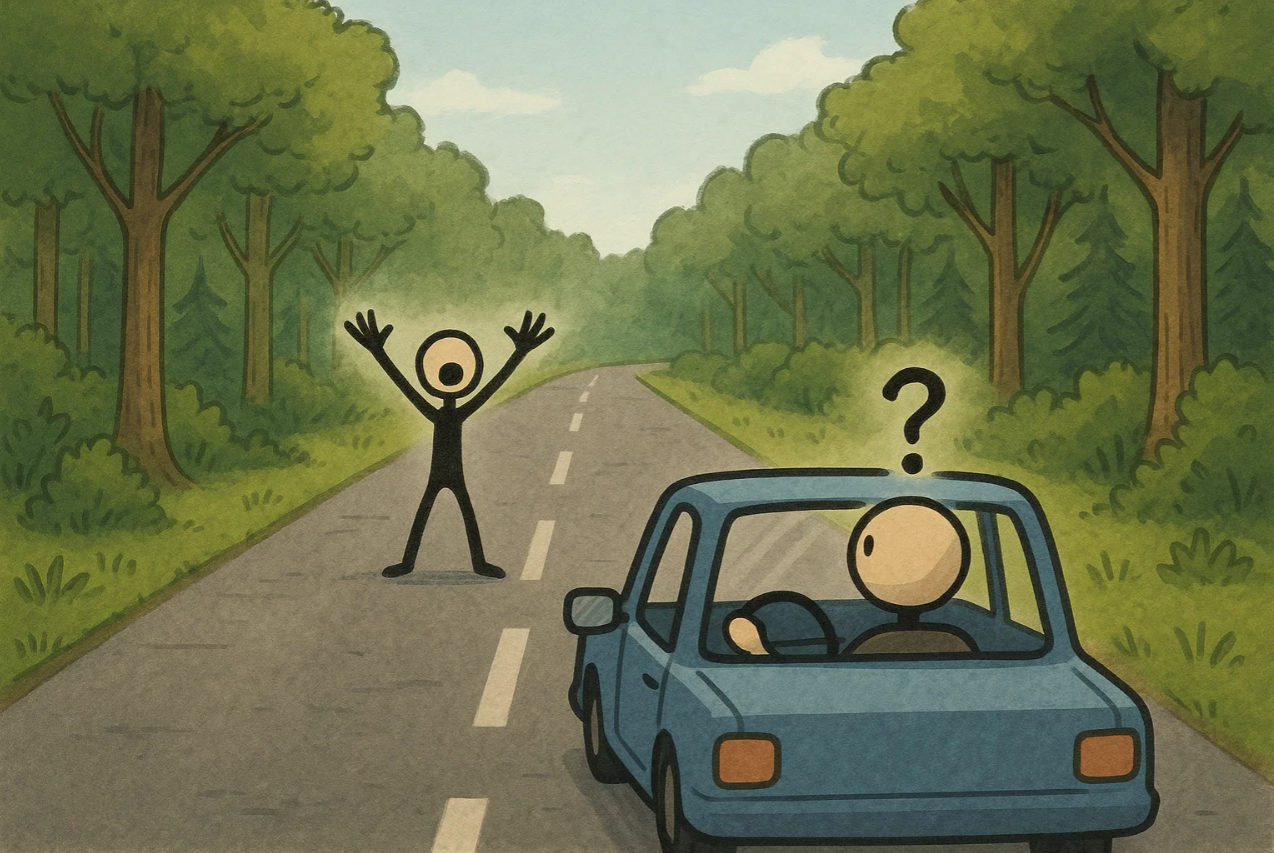}%
  \label{fig:stim2}}
\hfil
\subfloat[Unclaimed wallet]{%
  \includegraphics[height=0.156\textheight,keepaspectratio]{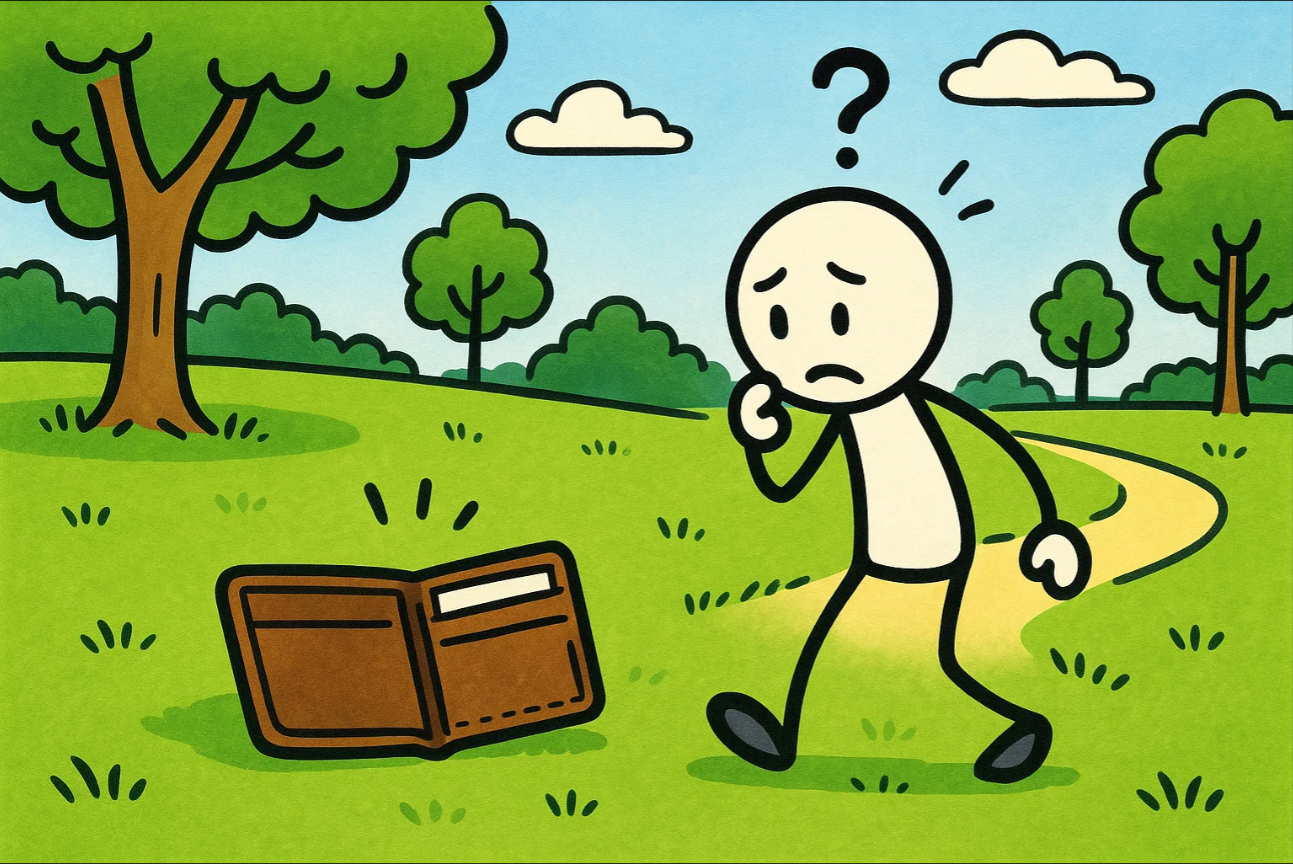}%
  \label{fig:stim3}}
\caption{Exemplar pictorial moral dilemmas used as stimuli in the study. The complete textual descriptions of 13 dilemmas are provided in Supplementary Table~S1.}
\label{fig:stimuli_strip3}
\end{figure*}

We used a performance-based moral decision-making task with 13 \emph{pictorial} dilemmas \textbf{(refer Fig. \ref{fig:stimuli_strip3} for instances of stimulus used)}, each showing real-life trade-offs \textbf{(see \textit{Supplementary Table S1 for more details})}. Pictorial vignettes reduce literacy demands, minimise framing bias, and increase ecological validity while remaining culturally neutral. Stick figures with no demographic cues ensured neutrality, and a focal figure with \emph{question marks} cued role-taking. The scenarios targeted prosocial orientation, perspective taking, reflective judgement, and practical wisdom. This task was preregistered on the Open Science Framework on October 28, 2024 (https://doi.org/10.17605/OSF.IO/TSM24
)

\subsection{Procedure: Think-Aloud, Decision, and Confidence Rating}
For each vignette, participants answered: \emph{“Imagine you are the figure with question marks. What would you do?”} They \emph{thought aloud} while reasoning, following protocol-analysis guidance \cite{ericsson1980}, then rated \emph{“How wise was your decision?”} on a scale of \textit{1 to 10}. Think-alouds capture context-sensitive cognitive processing aligned with fuzzy models \cite{ericsson1993, lewis1993}. Across 13 items per person, this yielded dense samples of moral cognitive processing for component inference.

\subsection{Audio Acquisition and Transcription}
In silent environments verbal reports were recorded using a Sony ICD-PX470. All 1300 files (100 $\times$ 13) were verbatim transcribed and noise/artifacts cleaned \cite{pennebaker2003, upadhye2020}. In contrast to the usual NLP pipelines \cite{jurafsky2023}, we did not discard hedges, negations, discourse markers, since they are indicators of reflective and cautious reasoning.

\subsection{Linguistic Feature Extraction and Wisdom Component Modelling}

Transcripts were analysed for five wisdom components: PT, Ref, Pro, REA, and IH \cite{ardelt2003, brienza2018, glueck2019, grossmann2016, huynh2023, koenig2014, staudinger2011psychological, webster2003}. To capture the five components of wisdom in participants’ narratives, we developed lexicons of linguistic markers through a systematic, iterative process. First, all 100 participants’ open-ended responses were carefully reviewed to identify recurring expressions that signaled the relevant constructs. Candidate markers were coded inductively, highlighting self-reflective, perspective-taking, prosocial orientation, empathic, or epistemically humble phrases. Next, the lists were refined with reference to established theoretical definitions of each construct in the wisdom literature, ensuring that the resulting lexicons were both empirically grounded in participant language and conceptually consistent with prior scholarship.

For \emph{Reflectiveness}, exemplars included phrases such as “I think,” “I realized,” “upon reflection,” “I thought about,” and “I came to understand.” For \emph{Perspective-Taking}, markers included, “if I were in that situation,” “I tried to see,” “from another point of view,” “I imagined myself,” and “I tried to understand.” \emph{Prosocial Orientation} was reflected in action-oriented responses that emphasized helping or norm-guided action, such as “try to return it,” “call the police,” “help the person,” “do the right thing,” and “save the child.” \emph{Reflective Empathetic Action} was identified through emotionally resonant expressions, including “I feel bad,” “I felt sorry,” “it breaks my heart,” “I would be sad,” and “my heart went out to them.” Finally, \emph{Intellectual Humility} was indexed by epistemic qualifiers and uncertainty markers such as “I might be wrong,” “I don’t know,” “I could be mistaken,” “maybe I am not sure,” and “it depends.” 

Categories were cross-checked with Linguistic Inquiry and Word Count (LIWC) resources \cite{pennebaker2003, tausczik2010} for coverage and balance.

\subsection{From Verbal Protocols to Fuzzy Inference}
Transcripts were tokenised, mapped to features, and fuzzified via KDE-based membership functions. These inputs fed a Mamdani inference engine \cite{mamdani1974} with 21 literature-based rules, yielding a \emph{wisdom score} (Attribute~A). Confidence ratings were normalized to $[0,1]$, segmented via KDE, and output as \emph{confidence} (Attribute~B). The final Z-number $Z=<A,B>$ thus pairs wisdom with perceived reliability, preserving interpretability and alignment with theory \cite{zadeh2011}.

\section{Attribute A: Wisdom Level Modeling}
In the proposed model, two complementary attributes are computed: 
\textbf{Attribute~A}, referred to as the \emph{Wisdom Score}, which represents a continuous quantitative estimate of wisdom; and 
\textbf{Attribute~B}, referred to as the \emph{Confidence Estimate}, which represents the associated degree of reliability of the score, expressed as a fuzzy membership value. 
These terms will be used consistently throughout the paper.

\subsection{Component-Level Feature Extraction}

To extract semantically meaningful features from participant responses to each dilemma, we adopted a marker-based linguistic framework aligned with established psychological constructs of wisdom. Each of the five components—Perspective-Taking (PT), Reflectiveness (Ref), Prosocial Orientation (PO), Reflective Empathetic Action (REA), and Intellectual Humility (IH)—was operationalized via curated word lists derived from prior studies and expert consensus.
\\

Let $x_{i,j}^{(k)}$ denote the feature score for participant $i$ on stimulus $j$ for component $k$, where $k \in \{1,\dots,5\}$. The extraction pipeline proceeds as follows:

\begin{enumerate}
    \item \textbf{Text Preprocessing}: Each written response is tokenized and lowercased. Stopwords are retained to preserve contextual integrity for markers (e.g., “not humble”).
    
    \item \textbf{Marker Matching}: For each component $k$, a dictionary $\mathcal{M}^{(k)}$ of psychologically validated markers is used. We count $n_{i,j}^{(k)}$, the number of tokens in response $(i,j)$ matching $\mathcal{M}^{(k)}$.
    
    \item \textbf{Length Normalization}: As shown in Eq. \eqref{eq:normalization} raw counts are normalized by total token count $N_{i,j}$ for fair comparison:
\begin{equation} \label{eq:normalization}
x_{i,j}^{(k)} = \min\left(1, \frac{n_{i,j}^{(k)}}{N_{i,j}}\right).
\end{equation}

    This bounds each feature within $[0,1]$, ensuring scale invariance.
    
\end{enumerate}

This marker-matching procedure provides interpretable, component-specific quantification of participant responses, forming the input for subsequent fuzzy modeling and Z-number construction.

\subsection{Z-number Construction and Membership Function Design for Each Component}
\label{sec:znumber_membership}

Each response-level component score \( x_{i,j}^{(k)} \in [0,1] \), where \( k \in \{\text{PT}, \text{REA}, \text{IH}, \text{PO}, \text{Ref} \} \), is fuzzified into qualitative labels (Low, Moderate, High) using component-specific membership functions.

To define these membership functions, we employed a data-driven segmentation procedure via Gaussian Kernel Density Estimation (KDE). For each component \(k\), KDE was applied to the distribution of all normalized values \( \{ x_{i,j}^{(k)} \} \), and local minima (valley points) in the estimated density function were used as fuzzy threshold boundaries \( \{t_1^{(k)}, \dots, t_6^{(k)}\} \). This avoids arbitrary binning and adapts the fuzzy structure to the actual empirical distribution.

As shown in Eq. \eqref{eq:kde} the KDE estimator \( \hat{f}_k(x) \) for each component is defined as:
\begin{equation} \label{eq:kde}
\hat{f}_k(x) = \frac{1}{n h} \sum_{j=1}^{n} K\left( \frac{x - x_j^{(k)}}{h} \right), \quad
K(u) = \frac{1}{\sqrt{2\pi}} e^{- \frac{1}{2} u^2},
\end{equation}
where \(h\) is the bandwidth, and \(x_j^{(k)}\) are normalized component scores.

Each component yielded six fuzzy breakpoints, which defined seven overlapping fuzzy regions using triangular or trapezoidal membership functions. These thresholds are summarized in \textbf{Table~\ref{tab:component_thresholds}}.

\begin{table}[h]
\scriptsize
\centering
\caption{KDE-Derived Thresholds for Component Membership Functions}
\label{tab:component_thresholds}
\begin{tabular}{lcccccc}
\toprule
\textbf{Component} & $t_1$ & $t_2$ & $t_3$ & $t_4$ & $t_5$ & $t_6$ \\
\midrule
PT & 0.10 & 0.22 & 0.38 & 0.52 & 0.66 & 0.81 \\
REA & 0.09 & 0.20 & 0.35 & 0.50 & 0.65 & 0.82 \\
IH & 0.12 & 0.25 & 0.40 & 0.54 & 0.68 & 0.84 \\
PO & 0.11 & 0.24 & 0.37 & 0.51 & 0.67 & 0.85 \\
Ref & 0.13 & 0.26 & 0.39 & 0.53 & 0.69 & 0.86 \\
\bottomrule
\end{tabular}
\end{table}

Using these thresholds, each \( x_{i,j}^{(k)} \) was mapped to fuzzy membership degrees \( \mu_{\text{low}}\) as shown in \cref{eq:membership-low}, \(\mu_{\text{mod}} \) as shown in \cref{eq:membership-mod}, \(\mu_{\text{high}} \) as shown in \cref{eq:membership-high}
, forming the qualitative component \( A_{i,j}^{(k)} \) in the Z-number \( Z_{i,j}^{(k)} = \langle A_{i,j}^{(k)}, B_{i,j} \rangle \), where \( A_{i,j}^{(k)} \in \{\text{Low}, \text{Moderate}, \text{High}\} \) denotes the linguistic wisdom category for component \(k\) (e.g., PT, REA), and \( B_{i,j} \in [0,1] \) is the normalized confidence score (Attribute B) for the participant \(i\) on stimulus \(j\). 

We define fuzzy membership functions for "Attribute A" as follows:
\begin{subequations}\label{eq:membership} % parent label
\begin{align}
\mu_{\text{low}}^{(k)}(x) &= \begin{cases}
   1 & x \le t_1^{(k)}\\
   \frac{t_2^{(k)}-x}{t_2^{(k)}-t_1^{(k)}} & t_1^{(k)}<x<t_2^{(k)}\\
   0 & \text{otherwise}
\end{cases} \label{eq:membership-low} \\[2ex]
\mu_{\text{mod}}^{(k)}(x) &= \begin{cases}
   0 & x \le t_2^{(k)} \text{ or } x \ge t_5^{(k)}\\
   \frac{x-t_2^{(k)}}{t_3^{(k)}-t_2^{(k)}} & t_2^{(k)}<x\le t_3^{(k)}\\
   1 & t_3^{(k)}<x<t_4^{(k)}\\
   \frac{t_5^{(k)}-x}{t_5^{(k)}-t_4^{(k)}} & t_4^{(k)}\le x < t_5^{(k)}
\end{cases} \label{eq:membership-mod} \\[2ex]
\mu_{\text{high}}^{(k)}(x) &= \begin{cases}
   0 & x \le t_5^{(k)}\\
   \frac{x-t_5^{(k)}}{t_6^{(k)}-t_5^{(k)}} & t_5^{(k)}<x<t_6^{(k)}\\
   1 & x \ge t_6^{(k)}
\end{cases} \label{eq:membership-high}
\end{align}
\end{subequations}

These piecewise functions ensure smooth, overlapping categorization while preserving interpretability. For each response, we compute the degree of membership in each category and assign the most likely one as in Eq.~\eqref{eq4}:
\begin{equation} \label{eq4}
    A_{i,j}^{(k)} = \arg\max_{c \in \{\text{low, mod, high}\}} \mu_c^{(k)}(x_{i,j}^{(k)}).
\end{equation}

\subsubsection*{Final Component-Level Z-number Assignment}

Each wisdom-related component \( k \in \{1,\dots,5\} \) is represented by a Z-number \( Z_{i,j}^{(k)} = \langle A_{i,j}^{(k)}, B_{i,j}^{(k)} \rangle \) for each participant \( i \in \{1,\dots,N\} \) and dilemma \( j \in \{1,\dots,J\} \). Here, \( A_{i,j}^{(k)} \) is a fuzzy membership vector over \{Low, Moderate, High\}, derived from KDE-based thresholds \( \{t_1^{(k)}, \dots, t_6^{(k)}\} \) (see Table~\ref{tab:component_thresholds}), while \( x_{i,j}^{(k)} \in [0,1] \) is the normalized response score.

The fuzzy membership vector is explicitly defined in Eq.~\eqref{eq5}:
\begin{equation} \label{eq5}
A_{i,j}^{(k)} = 
\left[
\mu_{\text{low}}^{(k)}(x_{i,j}),\ 
\mu_{\text{mod}}^{(k)}(x_{i,j}),\ 
\mu_{\text{high}}^{(k)}(x_{i,j})
\right],
\end{equation}
where \( \mu_{\cdot}^{(k)}(\cdot) \) are trapezoidal/triangular membership functions.

To obtain a participant-level score, we average the \( J \) response-level vectors as shown in Eq.~\eqref{eq6}:
\begin{equation} \label{eq6}
A_{i}^{(k)} = \frac{1}{J} \sum_{j=1}^{J} A_{i,j}^{(k)} =
\left[
\bar{\mu}_{\text{low}}^{(k)},
\bar{\mu}_{\text{mod}}^{(k)},
\bar{\mu}_{\text{high}}^{(k)}
\right].
\end{equation}
The dominant membership category is then identified according to Eq.~\eqref{eq7}
:
\begin{equation} \label{eq7}
A_{i}^{(k)} = \arg\max
\left[
\bar{\mu}_{\text{low}}^{(k)},
\bar{\mu}_{\text{mod}}^{(k)},
\bar{\mu}_{\text{high}}^{(k)}
\right].
\end{equation}

Thus, the final linguistic label for each component is extracted using Eq.~\eqref{eq8}:
\begin{equation} \label{eq8}
\text{Label}_i^{(k)} =
\arg\max_{\ell \in \{\text{low},\text{mod},\text{high}\}} \bar{\mu}_\ell^{(k)}.
\end{equation}

Each participant yields five Attribute~A values \( A_{i}^{(k)} \) and five labels \( \text{Label}_i^{(k)} \) across the wisdom components (PT, REA, IH, PO, Ref), as detailed in \textbf{Algorithm~\ref{alg:component_level_znumber}}.

\begin{algorithm}[H]
\small
\caption{Final Component-Level Attribute A Computation}
\label{alg:component_level_znumber}
\begin{algorithmic}[1]
\STATE \textbf{Input:} 
    Normalized scores \( x_{i,j}^{(k)} \in [0,1] \), KDE-derived thresholds \( \{t_1^{(k)},\dots,t_6^{(k)}\} \)
\FOR{each participant \( i = 1,\dots,N \)}
  \FOR{each component \( k = 1,\dots,5 \)}
    \STATE Initialize: \( \text{sum}_{\text{low}}, \text{sum}_{\text{mod}}, \text{sum}_{\text{high}} \gets 0 \)
    \FOR{each response \( j = 1,\dots,13 \)}
      \STATE Compute fuzzy memberships:
      \[
        A_{i,j}^{(k)} =
        \bigl[\mu_{\text{low}}^{(k)}(x_{i,j}^{(k)}),\;
              \mu_{\text{mod}}^{(k)}(x_{i,j}^{(k)}),\;
              \mu_{\text{high}}^{(k)}(x_{i,j}^{(k)})\bigr]
      \]
      \STATE Accumulate:
      \[
      \begin{aligned}
        \text{sum}_{\text{low}} &\mathrel{+}= \mu_{\text{low}}^{(k)}(x_{i,j}^{(k)}), \quad
        \text{sum}_{\text{mod}} \mathrel{+}= \mu_{\text{mod}}^{(k)}(x_{i,j}^{(k)}),\\
        \text{sum}_{\text{high}} &\mathrel{+}= \mu_{\text{high}}^{(k)}(x_{i,j}^{(k)})
      \end{aligned}
      \]
    \ENDFOR
    \STATE Compute average membership degrees:
    \[
    \bar{\mu}_{\text{low}}^{(k)} \gets \frac{\text{sum}_{\text{low}}}{13},\quad
    \bar{\mu}_{\text{mod}}^{(k)} \gets \frac{\text{sum}_{\text{mod}}}{13},\quad
    \bar{\mu}_{\text{high}}^{(k)} \gets \frac{\text{sum}_{\text{high}}}{13}
    \]
    \STATE Form aggregated fuzzy vector:
    \[
      A_i^{(k)} \gets \argmax \bigl[\bar{\mu}_{\text{low}}^{(k)},\ \bar{\mu}_{\text{mod}}^{(k)},\ \bar{\mu}_{\text{high}}^{(k)}\bigr]
    \]
    \STATE Assign linguistic label:
    \[
      \text{Label}_i^{(k)} \gets \argmax \bigl(A_i^{(k)}\bigr)
    \]
  \ENDFOR
\ENDFOR
\STATE \textbf{Output:} Final Component-level Attribute A value \( A_i^{(k)} \) and \( \text{Label}_i^{(k)} \)
\end{algorithmic}
\end{algorithm}

\section{Attribute B: Confidence Factor Modeling}

In line with Zadeh’s definition of Attribute B as an inherent measure of confidence, we operationalized B as participants’ self-reported certainty about the wisdom of their own decision (scale 1–10). Importantly, this instantiation does not treat B as a statistical reliability coefficient, but rather as a subjective confidence judgment—consistent with Zadeh’s allowance for self-reported or linguistically expressed certainty. Although subjective, these ratings capture how participants appraised their own choices, and were normalized prior to incorporation into the Z-number model.

\subsection{Normalization and Participant-Level Aggregation}
As shown in Eq. \eqref{norm_per} per-item ratings are linearly rescaled to $B_{i,j}\in[0,1]$:
\begin{equation} \label{norm_per}
B_{i,j}=\frac{r_{i,j}-1}{9},
\end{equation}
and averaged to obtain a participant-level confidence factor as shown in Eq. \eqref{avg},
\begin{equation} \label{avg}
B_i=\frac{1}{J}\sum_{j=1}^{J} B_{i,j},
\end{equation}
where $J=13$ dilemmas.

\subsection{Data-Driven Segmentation and Fuzzy Labeling}
We segment confidence using Gaussian KDE–based valley detection (bandwidth via Silverman’s rule \cite{silverman1986}; details in \textbf{\textit{(Supplementary Sec.~S2)}}. This yields seven linguistic categories
\{\textit{Perhaps, Possibly, Probably, Supposedly, Expectedly, Decisively, and Certainly}\}
represented by overlapping triangular/trapezoidal membership functions. Rather than the typical low/medium/high scheme, we aimed to reflect subtle epistemic gradations in confidence expression, which are central to the Z-number representation of uncertainty. The final label as shown in Eq. \eqref{fin_lab} is assigned by maximum membership:
\begin{equation} \label{fin_lab}
\ell_i^\ast=\arg\max_{\ell}\,\mu_\ell(B_i).
\end{equation}
The seven labels, their full piecewise definitions, numeric thresholds $(t_0\ldots t_8)$, the KDE plot, and the algorithmic pseudocode are provided in \textbf{\textit{Supplementary Sec.~S2}}.

\paragraph*{Note on Rule Base Placement}
The complete 21-rule base for Attribute~A is reported in \textit{\textbf{Supplementary Table S2 Section S3}}.

\section{Final Wisdom-Level Z-Number Inference}

We compute a unified wisdom score for each participant using a Mamdani-type fuzzy inference system (FIS) that integrates the five component-level fuzzy inputs (IH, PT, PO, REA, Ref) into a single \emph{Attribute~A} value and pairs it with \emph{Attribute~B} (confidence) to form a Z-number.

\subsection*{Fuzzy Inference Framework}
Each participant is characterized by five fuzzy inputs:
\[
\boldsymbol{\mu}_i = \left[ \mu^{\text{(IH)}},\ \mu^{\text{(PT)}},\ \mu^{\text{(Pro)}},\ \mu^{\text{(REA)}},\ \mu^{\text{(Ref)}} \right],
\]
where each \( \mu^{(k)} \in \{\text{Low},\ \text{Moderate},\ \text{High}\} \) represents the dominant linguistic label for the component \(k \in \{\text{IH, PT, Pro, REA, Ref}\}\) determined from their respective Attribute A scores.
A literature-grounded rule base maps combinations of these inputs to an output wisdom category. For example:
\[
\text{IF } \text{IH}=\text{High} \land \text{PT}=\text{High} \land \text{Ref}=\text{High} \Rightarrow \text{Wisdom}=\text{High}.
\]
Rule firing strength is computed using the minimum T-norm (see Eq.~\eqref{eq12}),
\begin{equation} \label{eq12}
\alpha_r=\min\bigl(\mu_{A_1}^{(r)},\mu_{A_2}^{(r)},\ldots\bigr),
\end{equation}
and consequents are clipped at $\alpha_r$. Aggregation across activated rules is carried out via the maximum S-norm (Eq.~\eqref{eq13}),
\begin{equation} \label{eq13}
\mu_{\text{out}}(x)=\max_{r}\,\mu_{C_r}^{\text{clipped}}(x),
\end{equation}
where $C_r\in\{\text{Low},\text{Moderate},\text{High}\}$. The output variable employs three overlapping triangular membership functions over $x\in[0,1]$ (\textit{Low}, \textit{Moderate}, \textit{High}) to preserve interpretability and smooth transitions.

\subsection*{Fuzzy Output Sets and Membership Functions}

The output fuzzy variable \( \mu^{\text{(Wisdom)}} \) is defined over the normalized universe of discourse \( x \in [0,1] \). We define three triangular membership functions corresponding to the linguistic categories of wisdom:

\begin{itemize}
    \item \textbf{Low Wisdom:} triangular function with base [0.0, 0.4], peak at 0.2
    \item \textbf{Moderate Wisdom:} triangular function with base [0.3, 0.7], peak at 0.5
    \item \textbf{High Wisdom:} triangular function with base [0.6, 1.0], peak at 0.8
\end{itemize}

Each fuzzy set is evaluated over its support — the interval where its membership is non-zero. To operationalize this, we discretize the universe \( x \in [0,1] \) using a step size \( \Delta x = 0.1 \). Thus, the complete set of points used in computations is:

\[
x \in \{0.0,\ 0.1,\ 0.2,\ 0.3,\ 0.4,\ 0.5,\ 0.6,\ 0.7,\ 0.8,\ 0.9,\ 1.0\}.
\]

However, for a given rule firing with consequent category \( C_r \), we only evaluate the output fuzzy set \( \mu_{C_r}(x) \) on its specific support. For example:

\begin{itemize}
    \item For \( C_r = \text{Low} \): \( x \in \{0.0,\ 0.1,\ 0.2,\ 0.3,\ 0.4\} \)
    \item For \( C_r = \text{Moderate} \): \( x \in \{0.3,\ 0.4,\ 0.5,\ 0.6,\ 0.7\} \)
    \item For \( C_r = \text{High} \): \( x \in \{0.6,\ 0.7,\ 0.8,\ 0.9,\ 1.0\} \)
\end{itemize}

Once a rule \( R_r \) is activated (i.e., has a non-zero firing strength), its associated fuzzy output set is clipped at the level of its firing strength as shown in below Eq. \eqref{firing}:

\begin{equation} \label{firing}
\mu_{C_r}^{\text{clipped}}(x) = \min\left( \mu_{C_r}(x),\ \text{FiringStrength}_r \right).
\end{equation}

This clipping restricts the fuzzy set to a maximum height equal to the degree of truth of its antecedent — i.e., the rule's firing strength. Only the fuzzy sets associated with activated rules are evaluated and clipped.

\subsection*{Aggregation of Clipped Output Sets}

After identifying and clipping the fuzzy sets corresponding to the fired rules, we aggregate these outputs Eq. \eqref{aggre} using the fuzzy union (maximum) S-norm operator:

\begin{equation} \label{aggre}
\mu_{\text{output}}(x) =
\max \Bigg(
\begin{aligned}
   &\mu_{\text{Low}}^{\text{clipped}}(x), \\
   &\mu_{\text{Moderate}}^{\text{clipped}}(x), \\
   &\mu_{\text{High}}^{\text{clipped}}(x)
\end{aligned}
\Bigg), \quad x \in [0,1].
\end{equation}

However, in practice, only the fuzzy sets corresponding to fired rules are included in the aggregation. For instance, if two rules are activated — one for “Moderate Wisdom” with firing strength 0.6 and one for “High Wisdom” with strength 0.7 — then the aggregated fuzzy set is computed as shown in Eq. \eqref{agg}:

\begin{equation} \label{agg}
\mu_{\text{output}}(x) = \max \left( \mu_{\text{Moderate}}^{\text{clipped}}(x),\ \mu_{\text{High}}^{\text{clipped}}(x) \right).
\end{equation}

Each \( \mu_{C}^{\text{clipped}}(x) \) is evaluated only over the support of its corresponding fuzzy category. The domain \( x \in [0,1] \) is uniformly discretized, typically using a step size of \( \Delta x = 0.1 \), i.e., \( x \in \{0.0, 0.1, \ldots, 1.0\} \).

\subsection*{Defuzzification and Z-Number Construction}
The crisp wisdom score is obtained by centroid defuzzification as shown in Eq. \eqref{def}:
\begin{equation} \label{def}
A_i^{\text{(final)}}=\frac{\sum_x x\,\mu_{\text{out}}(x)}{\sum_x \mu_{\text{out}}(x)}.
\end{equation}
The participant-level Z-number is computed as Eq. \eqref{z_num}
\begin{equation} \label{z_num}
Z_i=\langle A_i^{\text{(final)}},\,B_i\rangle,
\end{equation}
with $B_i\in[0,1]$ from Attribute~B.

\subsection*{Label Assignment}
As shown in Eq. \eqref{label_as} a categorical wisdom label accompanies $A_i^{\text{(final)}}$ by maximum membership over \{\textit{Low}, \textit{Moderate}, \textit{High}\}:
\begin{equation} \label{label_as}
\text{Label}_i=\arg\max_{\ell\in\{\text{Low,\,Moderate,\,High}\}}\ \mu_{\ell}\bigl(A_i^{\text{(final)}}\bigr).
\end{equation}

\section{Construct Validity}

To evaluate construct validity of the Z-number wisdom model, we tested convergent and divergent correlations of Attribute~A. Convergent validity was assessed against established wisdom measures (SDWISE, SAWS, Perspective Taking, Empathic Concern), while divergent validity was evaluated with unrelated traits (HEXACO domains) and fluid intelligence (Raven’s Progressive Matrices).

Because most variables deviated from normality (Shapiro--Wilk, $p < .05$), nonparametric Spearman correlations were used. Significance was set at $\alpha = .05$, with 95\% bias-corrected accelerated (BCa) confidence intervals estimated from 10{,}000 bootstrap resamples, and FDR correction applied across tests.

As shown in \textbf{Table~\ref{tab:conv_div_validity}}, Attribute~A correlated positively with SDWISE ($\rho = .222$, $p = .026$, 95\% CI [.012, .406]), SAWS ($\rho = .203$, $p = .043$, CI [--.029, .395]), and Perspective Taking ($\rho = .220$, $p = .028$, CI [.046, .372]), suggesting modest convergence with wisdom-related constructs. No significant association was observed with Empathic Concern ($\rho = .115$, $p = .257$, CI [--.109, .321]). Divergent measures (HEXACO traits, RPM) showed negligible or nonsignificant correlations, consistent with divergent validity. Given their small magnitude, these associations serve as proof-of-concept consistency checks in this sample (N=100) rather than definitive validation.

\begin{table}[!t]
\footnotesize
\caption{Convergent and Divergent Validity of Attribute A in the Z-Number Wisdom Model ($N{=}100$)}
\label{tab:conv_div_validity}
\centering
\setlength{\tabcolsep}{4.5pt}
\renewcommand{\arraystretch}{1.15}
\resizebox{\columnwidth}{!}{%
\begin{tabular}{|l|c|c|c|}
\hline
\textbf{External Measure} & \textbf{$\rho$} & \textbf{p-value} & \textbf{95\% CI (BCa)} \\
\hline
\multicolumn{4}{|c|}{\textit{Convergent Measures}} \\
\hline
San Diego Wisdom Scale (SDWISE) & 0.222$^{*}$ & 0.026 & [0.012, 0.406] \\
Self-Assessed Wisdom Scale (SAWS) & 0.203$^{*}$ & 0.043 & [-0.029, 0.395] \\
Perspective Taking (PT) & 0.220$^{*}$ & 0.028 & [0.046, 0.372] \\
Empathic Concern (EC) & 0.115 & 0.257 & [-0.109, 0.321] \\
\hline
\multicolumn{4}{|c|}{\textit{Divergent Measures}} \\
\hline
HEXACO – Honesty-Humility & -0.056 & 0.580 & [-0.201, 0.092] \\
HEXACO – Emotionality & -0.034 & 0.738 & [-0.230, 0.177] \\
HEXACO – Extraversion & -0.310$^{**}$ & 0.002 & [-0.468, -0.122] \\
HEXACO – Agreeableness & 0.093 & 0.357 & [-0.144, 0.316] \\
HEXACO – Conscientiousness & 0.101 & 0.320 & [-0.117, 0.279] \\
HEXACO – Openness to Experience & 0.143 & 0.155 & [-0.098, 0.343] \\
Raven’s Progressive Matrices (RPM) & 0.025 & 0.803 & [-0.153, 0.196] \\
\hline
\end{tabular}%
}
\par\smallskip
\raggedright\footnotesize $^{**}p<0.01$; $^{*}p<0.05$ (two-tailed).
\end{table}

\section{Results and Discussion}

In this study, we present a  mathematically rigorous model of wisdom, operationalized as \textit{Attribute~A} (Wisdom Score) and \textit{Attribute~B} (Confidence), on the basis of both behavioral decision-making data and self-perceived ratings of confidence as well as fuzzy logic-based membership functions. The methodology connects computational modeling and  psychological assessment, allowing one to quantify a construct that is traditionally considered an abstract, qualitative concept. Besides internal validity of the model demonstrated here, the results identify interpretable patterns of wisdom-related attributes that may guide the design of artificial intelligence as well as human factors research.

\begin{figure}[htbp]
    \centering
    \includegraphics[width=0.85\linewidth]{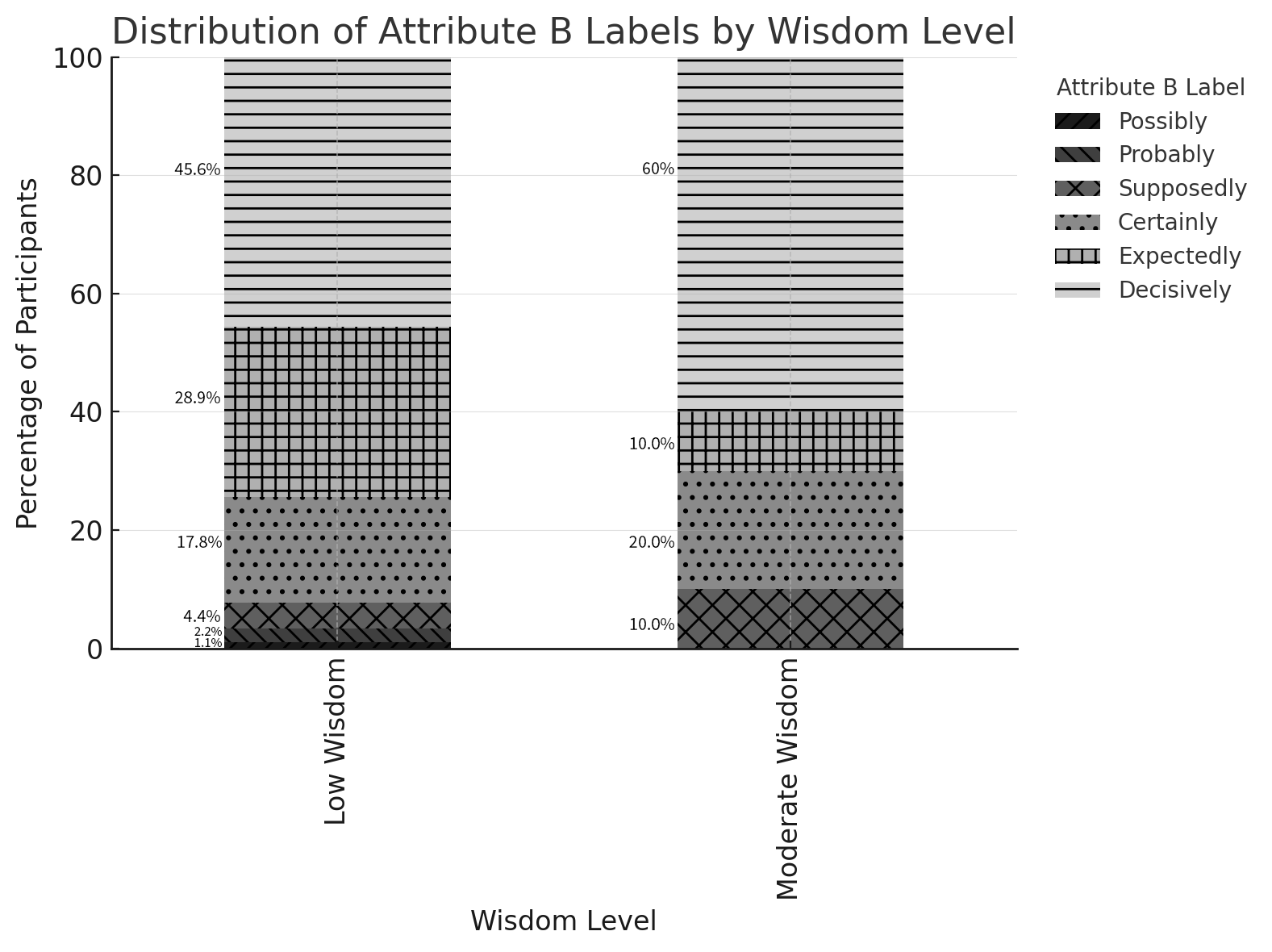}
    \caption{Percentage breakdown of the participants on fuzzy label of \textit{Attribute~B} by \textit{Wisdom Level} cluster. The order of the categories is smallest to largest average proportion by level of wisdom.}
    \label{fig:wisdom_distribution}
\end{figure}

\textbf{Figure~\ref{fig:wisdom_distribution}} illustrates the distribution of subjects in proportions on the basis of categories of \textit{Attribute~B}, divided into two groups, based on low and moderate wisdom scores. None of the participants reached the threshold of ``High Wisdom'' according to current model criteria, which, though indicative of the relative conservatism of such modeling standards, is equally suggestive of high selectivity in defining the actual measures of high-level wisdom. The findings demonstrate that significantly larger frequency of \textit{Decisively} can be found in the moderate wisdom participants ($+14.44\%$ difference), but \textit{Expectedly} appeared most frequently among the low wisdom group ($+18.89\%$ difference). These results suggest that patterns of decision-making confidence: measured in terms of the fuzzified \textit{Attribute~B}, are associated with higher-order characterisations of wisdom.

\begin{figure}[!t]
    \centering
    \includegraphics[width=0.95\columnwidth]{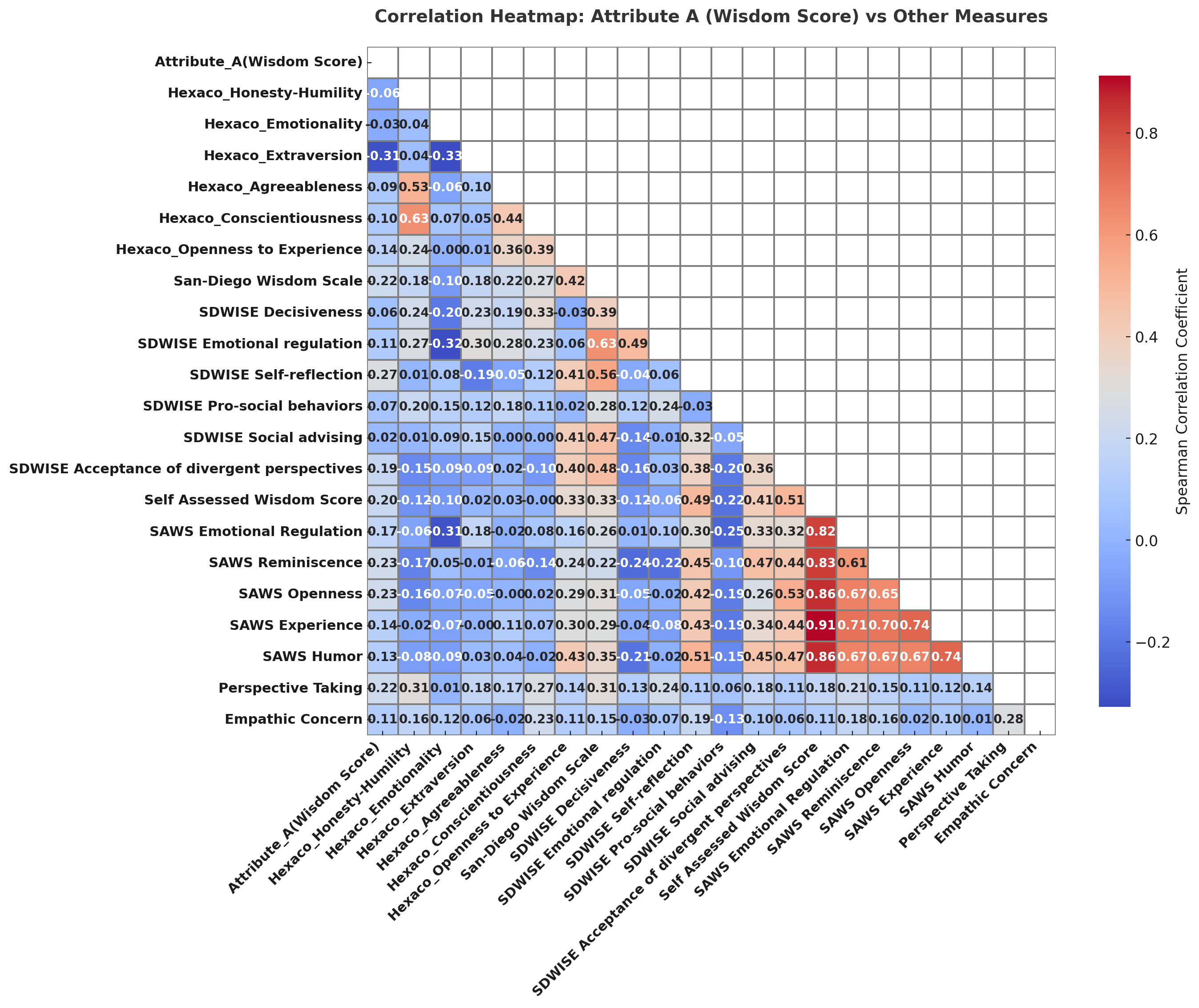}
    \caption{Upper-triangle Spearman correlation heatmap between \textbf{Attribute\_A (Wisdom Score)} and external measures of wisdom, personality, and empathy.}
    \label{fig:wisdom_heatmap}
\end{figure}

We analyzed how \textbf{\textit{Attribute~A} (Wisdom Score)} from our model, relates to established psychological measures (convergent and divergent validity). As shown in \textbf{Figure~\ref{fig:wisdom_heatmap}}, the observed correlations are consistent with expected patterns of convergent and divergent associations: \textit{Attribute~A} relates modestly to established wisdom scales and weakly to unrelated personality facets. This trend supports the theoretical validity of the fuzzy wisdom model and its non-dependence on non-target constructs.

\begin{figure}[htbp]
    \centering
    \includegraphics[width=0.89\linewidth]{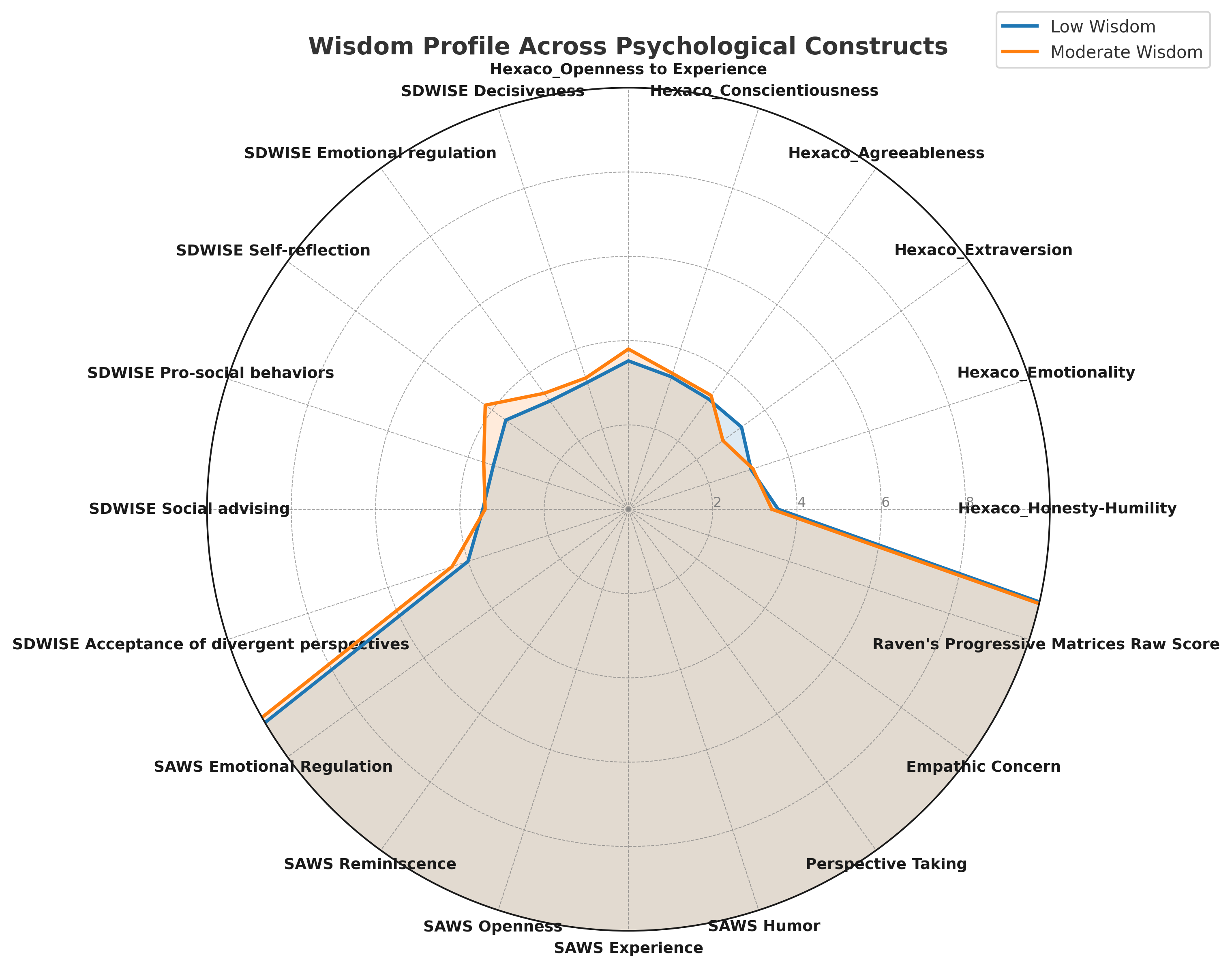}
    \caption{Radar plot of average scores across wisdom-related subcomponents (SDWISE, SAWS, empathy measures, HEXACO Honesty-Humility and Openness) for ``Low'' and ``Moderate'' wisdom groups.}
    \label{fig:wisdom_radar}
\end{figure}

To offer a more comprehensive overview of the construct of wisdom, \textbf{figure~\ref{fig:wisdom_radar}} illustrates a comparison between the profiles regarding the multidimensional assessment of ``Low Wisdom'' and ``Moderate Wisdom'' groups on six key \textbf{SDWISE} subcomponents (Decisiveness, Emotional Regulation, Self-Reflection, Pro-Social Behaviors, Social Advising, Acceptance of Divergent Perspectives), five \textbf{SAWS} facets (Emotional Regulation, Reminiscence, Openness, Experience, Humor), six key \textbf{HEXACO} subcomponents (Openness to Experience, Conscientiousness, Agreeableness, Extraversion, Emotionality, Honesty Humility), Perspective Taking and Empathic Concern, and Raven's Progressive Matrices. It shows that moderate wisdom participants are more likely to score high on most facets, especially in Emotional Regulation, Acceptance of Divergent Perspectives and Perspective Taking. 

We examined the demographic and group differences in modeled scores of wisdom. \textbf{Table~\ref{tab:gender_diff}} shows a summary of the results of a Mann-Whitney U test, showing that there is a statistically significant difference in gender in the sense that the females have a higher mean rank in \textit{Attribute~A} than males at the significance level of $p = 0.007$. This implies that the represented construct can be conditioned by gender-related variations in decision-making or self-concepts related to wisdom-related judgments, a factor that should be examined in cross-cultural and longitudinal settings.

\begin{table}[!ht]
\caption{Mann–Whitney U Test for Gender Differences in Attribute A}
\label{tab:gender_diff}
\centering
\renewcommand{\arraystretch}{1.2}
\setlength{\tabcolsep}{6pt}
\footnotesize
\begin{tabular}{|l|c|c|c|c|c|}
\hline
\textbf{Gender} & \textbf{N} & \textbf{Mean Rank} & \textbf{Sum of Ranks} & \textbf{U} & \textbf{p-value} \\
\hline
Male   & 47 & 45.96 & 2160.00 & 1032.0 & 0.007 \\
Female & 53 & 54.53 & 2890.00 & 1032.0 & 0.007 \\
\hline
\multicolumn{6}{l}{\footnotesize Mann–Whitney Z = -2.719, two-tailed test.} \\
\end{tabular}
\end{table}

\textbf{Table~\ref{tab:wisdom_kw}} summarises the Kruskal--Wallis test of the mean differences between wisdom levels on the \textit{Attribute~A}. The variation between the ``Low'' and ``Moderate'' categories of wisdom is very high ($p < 0.001$) indicating a considerable distance between mean ranks (45.50 and 95.50). This provides preliminary evidence that the model can discriminate between these groups; however, the result should be viewed as illustrative given the sample size and imbalance. 

\begin{table}[!ht]
\scriptsize
\caption{Kruskal--Wallis Test for Differences in Attribute~A Across Wisdom Levels (N=100)}
\label{tab:wisdom_kw}
\centering
\renewcommand{\arraystretch}{1.2}
\setlength{\tabcolsep}{5pt}
\footnotesize
\begin{tabular}{|l|c|c|}
\hline
\textbf{Wisdom Level} & \textbf{N} & \textbf{Mean Rank} \\
\hline
Low      & 90 & 45.50 \\
Moderate & 10 & 95.50 \\
\hline
\end{tabular}

\vspace{2mm}

\begin{tabular}{|l|c|}
\hline
\textbf{Test Statistic} & \textbf{Value} \\
\hline
Chi-Square ($H$) & 90.91 \\
df & 1 \\
Asymp. Sig. (2-tailed) & $<0.001$ \\
\hline
\end{tabular}
\end{table}

The sample was composed mostly of college students, few working adults, and a few middle school teenagers (ages 10 to 39 years (mean age 23)). Since wisdom has been related to life experience, the type of demographic profile is bound to contribute to lower measured wisdom levels. Moreover it was specifically designed that the model would be more specific in detecting high wisdom, through the fuzzy classification engine in the model, that consists of 21 Mamdani-type rules that are rigorously derived out of prior wisdom literature: this was to ensure that scores were not inflated. This conservative criterion guarantees that only those participants who fulfill several, literature-supported criteria would obtain a ``High Wisdom'' category. The lack of such cases in the sample at hand emphasizes the greater discriminative potential of the model than insufficiency and creates the impression that future research with older, more experience-diverse scholars could result in the high wisdom category.

\section{Conclusion and Future Work}

Based on empirical and theoretical data, this study proposes a new approach to the computational modelling of human wise decision-making as a fuzzy attribute-based system. This Z-number method permits one to convert multidimensional evidence (verbatims from think aloud task, measurements of SDWISE, SAWS, HEXACO, Perspective Taking, Empathic Concern, Raven's Progressive Matrices, and self-perceived confidence of wise decision) into an opaque \emph{Wisdom Score} (Attribute~A) coupled with a confidence estimate (Attribute~B). The key wisdom constructs were encoded by a literature-based set of 21 rules, and membership functions were data-calibrated using Gaussian KDE and valley detection, although calibration was based on the present sample, and it preserved theoretical fidelity as well as reproducibility.

Empirically, Attribute~A exhibited \emph{small-to-moderate} relationships with theoretically related subscales and insignificant relationships with unrelated traits, supporting convergent and discriminating validity. The profiling procedure identified primarily ‘Low’ and ‘Moderate’ wisdom categories, with no cases classified as ‘High.’ Although the observed convergent correlations were small in magnitude ($\rho \approx .20$--$.22$), they reached statistical significance due to the sample size ($N = 100$; the minimum detectable correlation at $\alpha = .05$, two-tailed, is $|\rho| \approx .197$). This indicates that the associations, while modest, are unlikely to be spurious, but they should be interpreted as reflecting limited shared variance ($\sim 4$--$5\%$) rather than a strong overlap.
Although simulations show that the system can detect such profiles given the right patterns of input, no case with “High Wisdom” was observed, thus indicating conservative thresholds rather than a structural ceiling of the system.
Although our empirical findings are modest, they primarily illustrate the feasibility of the model. The central advance lies in the computational framework, which integrates fuzzy rule-based reasoning, kernel density estimation (KDE)-based membership calibration, and Z-number uncertainty handling.

\textbf{Limitations.} The theory-based 21-rule base may not capture all possible combinations of antecedents, and the KDE-based memberships may be sample-specific. The sensitivity to parameter perturbations has not yet been measured. Although the fuzzy inference system is interpretable at the rule and pathway levels, the parameterization of membership functions (e.g., kernel bandwidths, valley detection criteria, and classification thresholds) introduces a layer of technical opacity. Although mathematically well-defined, such parameters may not always map transparently onto psychological constructs. In addition, the present sample was skewed toward younger participants (ages 10--39, $M \approx 23$), whereas many theories of wisdom emphasize later-life development. Although wisdom is often conceptualized as emerging later in life, our sample included adolescents (aged 10–17). 
This inclusion may limit construct validity, as younger participants have had fewer opportunities for life experience. 
At the same time, it provided a chance to examine developmental variability and what may be considered “nascent wisdom” or wisdom-related dispositions in youth. Indeed, prior work suggests that the seeds of wisdom can already be observed during adolescence, particularly in the form of reflective thinking and perspective-taking, even if the full integration of these components matures later in adulthood \cite{staudinger2011psychological, staudinger1999}.
Future work should replicate the fuzzy wisdom model in older and more experience-rich populations to better align with the theoretical accounts of wisdom.
Thus, the generalizability of the fuzzy wisdom model to older adult populations remains to be established. Moreover, all participants were recruited from a single national and cultural background (i.e., India). Therefore, future studies should evaluate the robustness of the model across diverse cultural contexts. The current findings should be contextualized in the culture of India, where all study participants were recruited. Although it is common to discuss the theoretical aspects of wisdom (e.g., perspective-taking, intellectual humility) across different cultures, their linguistic and behavioral manifestations can differ. Therefore, the currently available lexicons and calibrations of models containing Indian participant data are likely not directly applicable to other cultural contexts. Future studies should expand the framework to a cross-cultural study to determine universal versus culture-specific wisdom indicators. In addition, because our fuzzy linguistic labels are novel, replication may require adapting this terminology; future work should test whether standard categories yield similar results. 

Despite these limitations, our findings offer a \textbf{principled starting point} for assessing computational wisdom. Future research that applies the fuzzy wisdom model across broader age ranges and cultural contexts will not only test its generalizability but also provide an opportunity to refine the rule base and the membership functions. In this sense, the current study provides a foundation for cross-cultural and lifespan-sensitive approaches to measuring wisdom.

\textbf{Future work.} We shall (i) extend and expert-validate the rule base to maximize coverage; (ii) cross-validate, bootstrap, and formally sensitivity-analyze memberships and rule-weights; (iii) sample more widely in terms of age, culture, and practice settings; (iv) benchmark classification thresholds on simulated and case-based high-wisdom exemplars; (v) triangulate against self-report by both behavior measures and observer ratings; and (vi) catalogue a prospective follow-up to track within-subject change. Bridging computational intelligence and psychological measurement theory, this study proposes a framework that can be replicated, extended, and explained, that can be used to measure wisdom. We believe that this model will contribute to research on computational psychology and will also be applicable to school, organizational, and clinical contexts where the cultivation and assessment of wisdom is of paramount importance.

\section*{Funding}
This research received no external funding.

\section*{Author Contributions}

A.S. provided the conceptual foundation for the study, contributing expertise on wisdom theories and guiding the development of the assessment framework. 
R.B. contributed technical expertise on Z-number theory, fuzzy modeling, and validation procedures. 
S.K. carried out the implementation, coding, and data analysis and drafted the manuscript. 
All authors collaborated in the design of the stimuli, provided critical feedback during the development of the methodology, contributed to manuscript revisions, and approved the final version for submission.

\section*{Data Availability}
The datasets generated and analyzed during the current study are not publicly archived but are available from the corresponding author upon reasonable request.

\section*{Competing Interests}
No, I declare that the authors have no competing interests as defined by Springer, or other interests that might be perceived to influence the results and/or discussion reported in this paper.

\clearpage
% Optional divider page (keeps main layout untouched)

\addcontentsline{toc}{section}{Supplementary Material} % harmless if no ToC

% Insert the *compiled* supplement as-is, full pages
\includepdf[pages=-,pagecommand={},linktodoc=true]{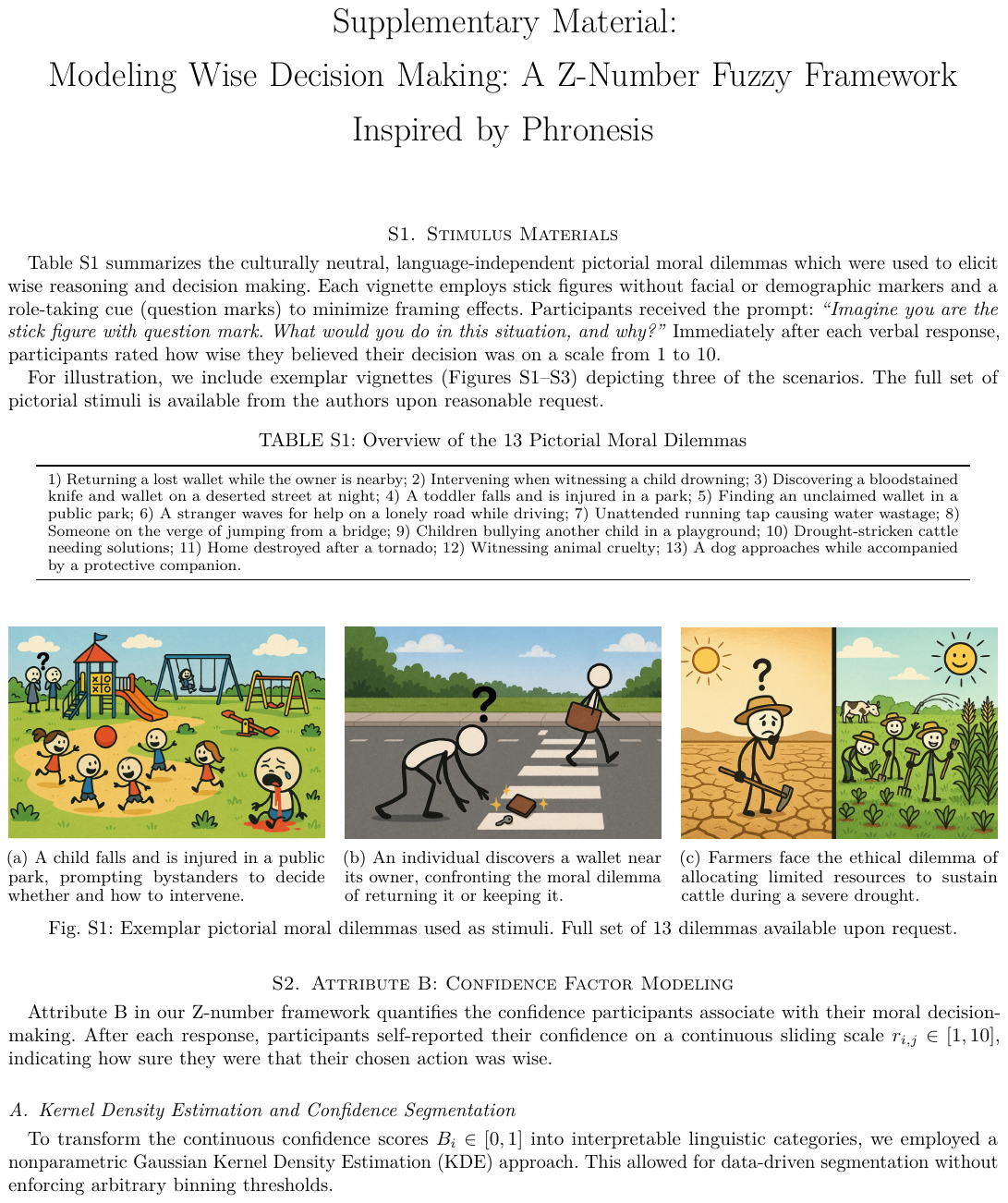}


\begin{thebibliography}{1}
\bibliographystyle{IEEEtran}

\bibitem{ardelt2003}
M. Ardelt, "Empirical assessment of a three-dimensional wisdom scale," \textit{Research on Aging}, vol. 25, no. 3, pp. 275--324, 2003.

\bibitem{ardelt2006}
Carstensen, L. L., Mikels, J. A., Mather, M., Birren, J. E., \& Schaie, K. W. (2006). Handbook of the psychology of aging.

\bibitem{glueck2019}
Glück, J., König, S., Naschenweng, K., Redzanowski, U., Dorner, L., Straßer, I., \& Wiedermann, W. (2013). How to measure wisdom: Content, reliability, and validity of five measures. \textit{Frontiers in Psychology, 4}, 405.


\bibitem{jeste2017}
Thomas, M. L., Bangen, K. J., Palmer, B. W., Martin, A. S., Avanzino, J. A., Depp, C. A., Glorioso, D., Daly, R. E., \& Jeste, D. V. (2019).  
A new scale for assessing wisdom based on common domains and a neurobiological model: The San Diego Wisdom Scale (SDWISE).  
\textit{Journal of Psychiatric Research, 108}, 40--47. Elsevier.


\bibitem{staudinger2011psychological}
U. M. Staudinger and J. Glück, "Psychological wisdom research: Commonalities and differences in a growing field," \textit{Annual Review of Psychology}, vol. 62, pp. 215--241, 2011.


\bibitem{staudinger1999}
U.~M.~Staudinger, ``Older and wiser? Integrating results on the relationship between age and wisdom-related performance,'' 
\emph{International Journal of Behavioral Development}, vol.~23, no.~3, pp.~641--664, 1999.


\bibitem{sternberg1998}
R.~J.~Sternberg, ``A balance theory of wisdom,'' 
\emph{Review of General Psychology}, vol.~2, no.~4, pp.~347--365, 1998.


\bibitem{sternberg1990wisdom}
R. J. Sternberg, \textit{Wisdom: Its Nature, Origins, and Development}. Cambridge, U.K.: Cambridge Univ. Press, 1990.

\bibitem{abraham2005}
A. Abraham, “130: Rule-based expert systems,” in \textit{Handbook of Measuring System Design}, P. H. Sydenham and R. Thorn, Eds., vol. 8. Chichester, UK: John Wiley \& Sons, Ltd., 2005, pp. 909–919.


\bibitem{mamdani1974}
E. H. Mamdani, "Application of fuzzy algorithms for control of simple dynamic plant," \textit{Proceedings of the Institution of Electrical Engineers}, vol. 121, no. 12, pp. 1585--1588, 1974.

\bibitem{mendel2017}
W. W. Tan and T. W. Chua, “Uncertain rule-based fuzzy logic systems: introduction and new directions (Mendel, J. M.; 2001) [book review],” \textit{IEEE Computational Intelligence Magazine}, vol. 2, no. 1, pp. 72–73, 2007.


\bibitem{zadeh2011}
L. A. Zadeh, “A note on Z-numbers,” \textit{Information Sciences}, vol. 181, no. 14, pp. 2923–2932, 2011.




\bibitem{robinson2014}
R. S. Robinson, “Purposive sampling,” in \textit{Encyclopedia of Quality of Life and Well-Being Research}, Springer, 2014, pp. 5243–5245.


\bibitem{goodman1961}
L. A. Goodman, “Snowball sampling,” \textit{The Annals of Mathematical Statistics}, pp. 148–170, 1961.


\bibitem{helsinki}
World Medical Association, "World Medical Association Declaration of Helsinki: Ethical principles for medical research involving human subjects," \textit{JAMA}, vol. 310, no. 20, pp. 2191--2194, 2013, doi: 10.1001/jama.2013.281053.

\bibitem{jurafsky2023}
D. Jurafsky and J. H. Martin, \textit{Speech and Language Processing: An Introduction to Natural Language Processing, Computational Linguistics, and Speech Recognition}. Upper Saddle River, NJ, USA: Pearson/Prentice Hall, 2009.


\bibitem{pennebaker2003}
J. W. Pennebaker, M. E. Francis, and R. J. Booth, \textit{Linguistic Inquiry and Word Count: LIWC 2001}. Mahwah, NJ, USA: Lawrence Erlbaum Associates, 2001.


\bibitem{tausczik2010}
Y. R. Tausczik and J. W. Pennebaker, ``The psychological meaning of words: LIWC and computerized text analysis methods,'' \textit{Journal of Language and Social Psychology}, vol. 29, no. 1, pp. 24--54, 2010.



\bibitem{upadhye2020}
A. Upadhye, ``A comprehensive survey of text data cleaning techniques: Challenges, methods, and best practices,'' \textit{Journal of Scientific and Engineering Research}, vol. 7, no. 5, pp. 205--210, 2020.


\bibitem{silverman1986}
B. W. Silverman, \textit{Density Estimation for Statistics and Data Analysis}. Routledge, 2018.



\bibitem{brienza2018}
P. Brienza, F. Y. H. Kung, H. C. Santos, D. R. Bobocel, and I. Grossmann, 
“Wisdom, bias, and balance: Toward a process-sensitive measurement of wisdom-related cognition,” 
\textit{Journal of Personality and Social Psychology}, vol. 115, no. 6, pp. 1093--1116, 2018.


\bibitem{grossmann2016}
I. Grossmann, 
“Wisdom and how to cultivate it,” 
\textit{European Psychologist}, vol. 22, no. 4, pp. 233--246, 2017.


\bibitem{huynh2023}
A. C. Huynh and I. Grossmann, 
“A pathway for wisdom-focused education,” 
\textit{Journal of Moral Education}, vol. 49, no. 1, pp. 9--29, 2020.


\bibitem{koenig2014}
S. K{\"o}nig and J. Gl{\"u}ck, 
``Gratitude is with me all the time: How gratitude relates to wisdom,'' 
\textit{Journals of Gerontology Series B: Psychological Sciences and Social Sciences}, 
vol. 69, no. 5, pp. 655--666, 2014.




\bibitem{webster2003}
J. D. Webster, 
``An exploratory analysis of a self-assessed wisdom scale,'' 
\textit{Journal of Adult Development}, 
vol. 10, no. 1, pp. 13--22, 2003.


\bibitem{ericsson1980}
K. A. Ericsson and H. A. Simon, 
``Verbal reports as data,'' 
\textit{Psychological Review}, 
vol. 87, no. 3, p. 215, 1980.


\bibitem{ericsson1993}
K. A. Ericsson and H. A. Simon, 
\textit{Protocol Analysis: Verbal Reports as Data}, Rev. ed. 
Cambridge, MA: MIT Press, 1993.


\bibitem{lewis1993}
C. Lewis and J. Rieman, 
\textit{Task-Centered User Interface Design: A Practical Introduction}. 
Boulder, CO: University of Colorado, Boulder, 1993.


\bibitem{kant1785}
I. Kant, \textit{Groundwork of the Metaphysics of Morals}. 
Riga: Johann Friedrich Hartknoch, 1785.


\bibitem{aristotle350}
M. Pakaluk, \textit{Aristotle's Nicomachean Ethics: An Introduction}. 
Cambridge: Cambridge University Press, 2005.


\bibitem{kohlberg1969}
L. Kohlberg, ``Stage and sequence: The cognitive-developmental approach to socialization,'' 
in \textit{The First Half of the Chapter is a Revision of a Paper Prepared for the Social Science Research Council, Committee on Socialization and Social Structure, Conference on Moral Development, Arden House, Nov 1963}. 
New York: Garland Publishing, 1994.


\bibitem{gilligan1982}
C. Gilligan, \textit{In a Different Voice: Psychological Theory and Women's Development}. 
Cambridge, MA: Harvard University Press, 1993.


\bibitem{haidt2001}
J. Haidt, ``The emotional dog and its rational tail: A social intuitionist approach to moral judgment,'' 
\textit{Psychological Review}, vol. 108, no. 4, pp. 814--834, 2001.


\bibitem{greene2004}
J. D. Greene, ``The dual-process theory of moral judgment does not deny that people can make compromise judgments,'' 
\textit{Proceedings of the National Academy of Sciences}, vol. 120, no. 6, p. e2220396120, 2023.


\bibitem{kahneman2011}
D. Kahneman, \textit{Thinking, Fast and Slow}. Macmillan, 2011.


\bibitem{haidt2004}
J. Haidt and C. Joseph, ``Intuitive ethics: How innately prepared intuitions generate culturally variable virtues,'' \textit{Daedalus}, vol. 133, no. 4, pp. 55--66, 2004.


\bibitem{kosko1994}
B. Kosko, \textit{The New Science of Fuzzy Logic}. New York, NY: HarperCollins, 1994.

\bibitem{zadeh1965}
L. A. Zadeh, ``Fuzzy sets,'' \textit{Information and Control}, vol. 8, no. 3, pp. 338--353, 1965.

\bibitem{bentham1789}
J. Bentham, \textit{An Introduction to the Principles of Morals and Legislation}. London, UK: T. Payne, 1789.

\bibitem{mill1863}
J. S. Mill, \textit{Utilitarianism}. Cambridge, MA: Harvard University Press, 2011. (Originally published 1863)

\bibitem{malle2016}
B. F. Malle, ``Integrating robot ethics and machine morality: the study and design of moral competence in robots,'' \textit{Ethics and Information Technology}, vol. 18, no. 4, pp. 243--256, 2016.

\bibitem{sharma2024_hindi}
A. Sharma and A. Sharma, ``Hindi wisdom and insights for global leaders: Indian conceptualization of wisdom-emic perspective and lessons for global leadership,'' in \textit{Global Leadership and Wisdoms of the World}, pp. 49--64. Cheltenham, UK: Edward Elgar Publishing, 2024.

\bibitem{shweder1997}
R. A. Shweder, M. Mahapatra, and J. G. Miller, ``Culture and moral development,'' in \textit{The Emergence of Morality in Young Children}, pp. 1--83, 1987.

\bibitem{markus1991}
H. R. Markus and S. Kitayama, ``Culture and the self: Implications for cognition, emotion, and motivation,'' in \textit{College Student Development and Academic Life}, pp. 264--293. New York, NY: Routledge, 2014.

\bibitem{triandis1995}
H. C. Triandis, ``Individualism and collectivism: Past, present, and future,'' Oxford University Press, 2001.

\bibitem{peng1999}
K. Peng and R. E. Nisbett, ``Culture, dialectics, and reasoning about contradiction,'' \textit{American Psychologist}, vol. 54, no. 9, pp. 741--754, 1999.

\bibitem{spencer2004}
J. Spencer-Rodgers, H. C. Boucher, S. C. Mori, L. Wang, and K. Peng, ``The dialectical self-concept: Contradiction, change, and holism in East Asian cultures,'' \textit{Personality and Social Psychology Bulletin}, vol. 35, no. 1, pp. 29--44, Jan. 2009, doi: 10.1177/0146167208325772.

\bibitem{henrich2010}
J. Henrich, S. J. Heine, and A. Norenzayan, ``The weirdest people in the world?,'' \textit{Behavioral and Brain Sciences}, vol. 33, no. 2–3, pp. 61--83, 2010.

\bibitem{hoppe2012}
A. Hoppe and M. Tabacchi, ``Towards a modelization of the elusive concept of wisdom using fuzzy techniques,'' in \textit{Proc. 2012 Annual Meeting of the North American Fuzzy Information Processing Society (NAFIPS)}, pp. 1--5, 2012.

\bibitem{hein2022}
A. Hein, L. J. Meier, A. M. Buyx, and K. Diepold, ``A fuzzy-cognitive-maps approach to decision-making in medical ethics,'' in \textit{Proc. 2022 IEEE Int. Conf. on Fuzzy Systems (FUZZ-IEEE)}, pp. 1--8, 2022.


\bibitem{banerjee2022}
R. Banerjee, S. K. Pal, and J. K. Pal, ``A decade of the Z-numbers,'' \textit{IEEE Transactions on Fuzzy Systems}, vol. 30, no. 8, pp. 2800--2812, 2021.


\bibitem{liao2024}
H. Liao, F. Liu, Y. Xiao, Z. Wu, and E. K. Zavadskas, ``A survey on Z-number-based decision analysis methods and applications: What’s going on and how to go further?,'' \textit{Information Sciences}, vol. 663, p. 120234, 2024.

\bibitem{aghaei2021}
H. Aghaei, M. M. Aliabadi, F. Mollabahrami, and K. Najafi, ``Human reliability analysis in de-energization of power line using HEART in the context of Z-numbers,'' \textit{PLOS One}, vol. 16, no. 7, p. e0253827, 2021.

\bibitem{anjaria2022}
K. Anjaria, ``Knowledge derivation from Likert scale using Z-numbers,'' \textit{Information Sciences}, vol. 590, pp. 234--252, 2022.


\bibitem{chai2023}
J. Chai, Y. Su, and S. Lu, ``Linguistic Z-number preference relation for group decision-making and its application in digital transformation assessment of SMEs,'' \textit{Expert Systems with Applications}, vol. 213, p. 118749, 2023.

\bibitem{pearl1988}
J. Pearl, \textit{Probabilistic Reasoning in Intelligent Systems: Networks of Plausible Inference}. Amsterdam, Netherlands: Elsevier, 2014.

\bibitem{shafer1976}
G. Shafer, \textit{A Mathematical Theory of Evidence}. Princeton, NJ: Princeton University Press, 2020.


\bibitem{baltes2000}
P. B. Baltes and U. M. Staudinger, ``Wisdom: A metaheuristic (pragmatic) to orchestrate mind and virtue toward excellence,'' \textit{American Psychologist}, vol. 55, no. 1, pp. 122--136, 2000.



\bibitem{brienza2017}
I. Grossmann, H. Oakes, and H. C. Santos, ``Wise reasoning benefits from emodiversity, irrespective of emotional intensity,'' \textit{Journal of Experimental Psychology: General}, vol. 148, no. 5, pp. 805--823, 2019.

\bibitem{grossmann2020}
I. Grossmann, N. M. Weststrate, M. Ardelt, J. P. Brienza, M. Dong, M. Ferrari, M. A. Fournier, C. S. Hu, H. C. Nusbaum, and J. Vervaeke, ``The science of wisdom in a polarized world: Knowns and unknowns,'' \textit{Psychological Inquiry}, vol. 31, no. 2, pp. 103--133, 2020.

\bibitem{dehghani2008}
M. Dehghani, E. Tomai, K. Forbus, R. Iliev, and M. Klenk, ``Moraldm: A computational model of moral decision-making,'' in \textit{Proc. 30th Annu. Conf. Cognitive Science Society (CogSci)}, 2008.




\bibitem{van2019}
J. M. van Baar, L. J. Chang, and A. G. Sanfey, ``The computational and neural substrates of moral strategies in social decision-making,'' \textit{Nature Communications}, vol. 10, no. 1, p. 1483, 2019.


\bibitem{lockwood2025}
P. L. Lockwood, W. van den Bos, and J.-C. Dreher, ``Moral learning and decision-making across the lifespan,'' \textit{Annual Review of Psychology}, vol. 76, no. 1, pp. 475--500, 2025.


\bibitem{guss2018}
C. D. Güss, ``What is going through your mind? Thinking aloud as a method in cross-cultural psychology,'' \textit{Frontiers in Psychology}, vol. 9, p. 1292, 2018.

\bibitem{bajovic2021meta}
M. Bajovic and K. Rizzo, ``Meta-moral cognition: bridging the gap among adolescents’ moral thinking, moral emotions and moral actions,'' \textit{International Journal of Adolescence and Youth}, vol. 26, no. 1, pp. 1--11, 2021.

\bibitem{kosinski2013}
M. Kosinski, D. Stillwell, and T. Graepel, ``Private traits and attributes are predictable from digital records of human behavior,'' \textit{Proceedings of the National Academy of Sciences}, vol. 110, no. 15, pp. 5802--5805, 2013.

\bibitem{bwp} 
P. B. Baltes and U. M. Staudinger, ``Wisdom: A metaheuristic (pragmatic) to orchestrate mind and virtue toward excellence,'' \textit{American Psychologist}, vol. 55, no. 1, p. 122, 2000.

\bibitem{brwp} 
C. Mickler and U. M. Staudinger, ``Personal wisdom: validation and age-related differences of a performance measure,'' \textit{Psychology and Aging}, vol. 23, no. 4, p. 787, 2008.

\bibitem{brienza2018situated}
J. P. Brienza, F. Y. Kung, H. C. Santos, R. Bobocel, and I. Grossmann, ``Situated Wise Reasoning Scale,'' \textit{Journal of Personality and Social Psychology}, 2018.



\bibitem{blass2015moral}
J. Blass and K. Forbus, ``Moral decision-making by analogy: Generalizations versus exemplars,'' in \textit{Proceedings of the AAAI Conference on Artificial Intelligence}, vol. 29, no. 1, Feb. 2015.

\bibitem{bangen2013defining}
K.~J. Bangen, T.~W. Meeks, and D.~V. Jeste, ``Defining and assessing wisdom: A review of the literature,'' \emph{The American Journal of Geriatric Psychiatry}, vol.~21, no.~12, pp. 1254--1266, 2013, Elsevier.

\bibitem{zhang2023wisdom}
K.~Zhang, J.~Shi, F.~Wang, and M.~Ferrari, ``Wisdom: Meaning, structure, types, arguments, and future concerns,'' \emph{Current Psychology}, vol.~42, no.~18, pp. 15030--15051, 2023, Springer.

\bibitem{phillips2021four}
P.~J. Phillips, C.~A. Hahn, P.~C. Fontana, D.~A. Broniatowski, and M.~A. Przybocki, ``Four principles of explainable artificial intelligence,'' \emph{NIST Interagency Report}, 2021.
\bibitem{xu2025ai}
H.~Xu, Y.~Wang, Y.~Xun, R.~Shao, and Y.~Jiao, ``Artificial intelligence for clinical reasoning: the reliability challenge and path to evidence-based practice,'' \emph{QJM: An International Journal of Medicine}, p. hcaf114, 2025.

\bibitem{umoke2025governance}
C.~C. Umoke, S.~O. Nwangbo, and O.~A. Onwe, ``The governance of AI in education: Developing ethical policy frameworks for adaptive learning technologies,'' unpublished.

\bibitem{lindenmeyer2025trustworthy}
A.~Lindenmeyer, M.~Blattmann, S.~Franke, T.~Neumuth, and D.~Schneider, ``Towards trustworthy AI in healthcare: Epistemic uncertainty estimation for clinical decision support,'' \emph{Journal of Personalized Medicine}, vol.~15, no.~2, p.~58, 2025.


\end{thebibliography}
\end{document}

\maketitle

\thispagestyle{empty}   % first page
\pagestyle{empty}
%=====================================================================
\section{Stimulus Materials}

Table~\ref{tab:stimuli_list} summarizes the culturally neutral, language-independent pictorial moral dilemmas which were used to elicit wise reasoning and decision
making. Each vignette employs stick figures without facial or demographic markers and a role-taking cue (question marks) to minimize framing effects. Participants received the prompt: \emph{``Imagine you are the stick figure with question mark. What would you do in this situation, and why?''} Immediately after each verbal response, participants rated how wise they believed their decision was on a scale from 1 to 10.

For illustration, we include exemplar vignettes (Figures~S1–S3) depicting three of the scenarios. 
The full set of pictorial stimuli is available from the authors upon reasonable request.

\begin{table}[!htbp]  % allow here, top, bottom, page
\centering
\caption{Overview of the 13 Pictorial Moral Dilemmas}
\label{tab:stimuli_list}
\begin{tabular}{p{0.92\columnwidth}}
\toprule
1) Returning a lost wallet while the owner is nearby; 
2) Intervening when witnessing a child drowning; 
3) Discovering a bloodstained knife and wallet on a deserted street at night; 
4) A toddler falls and is injured in a park; 
5) Finding an unclaimed wallet in a public park; 
6) A stranger waves for help on a lonely road while driving; 
7) Unattended running tap causing water wastage; 
8) Someone on the verge of jumping from a bridge; 
9) Children bullying another child in a playground; 
10) Drought-stricken cattle needing solutions; 
11) Home destroyed after a tornado; 
12) Witnessing animal cruelty; 
13) A dog approaches while accompanied by a protective companion. \\
\bottomrule
\end{tabular}
\end{table}
\FloatBarrier

\begin{figure}[H]
\centering
\begin{subfigure}{0.32\textwidth}
    \includegraphics[width=\linewidth]{stim2.jpg}
    \caption{A child falls and is injured in a public park, prompting bystanders to decide whether and how to intervene.}
    \label{fig:stim1}
\end{subfigure}
\hfill
\begin{subfigure}{0.32\textwidth}
    \includegraphics[width=\linewidth]{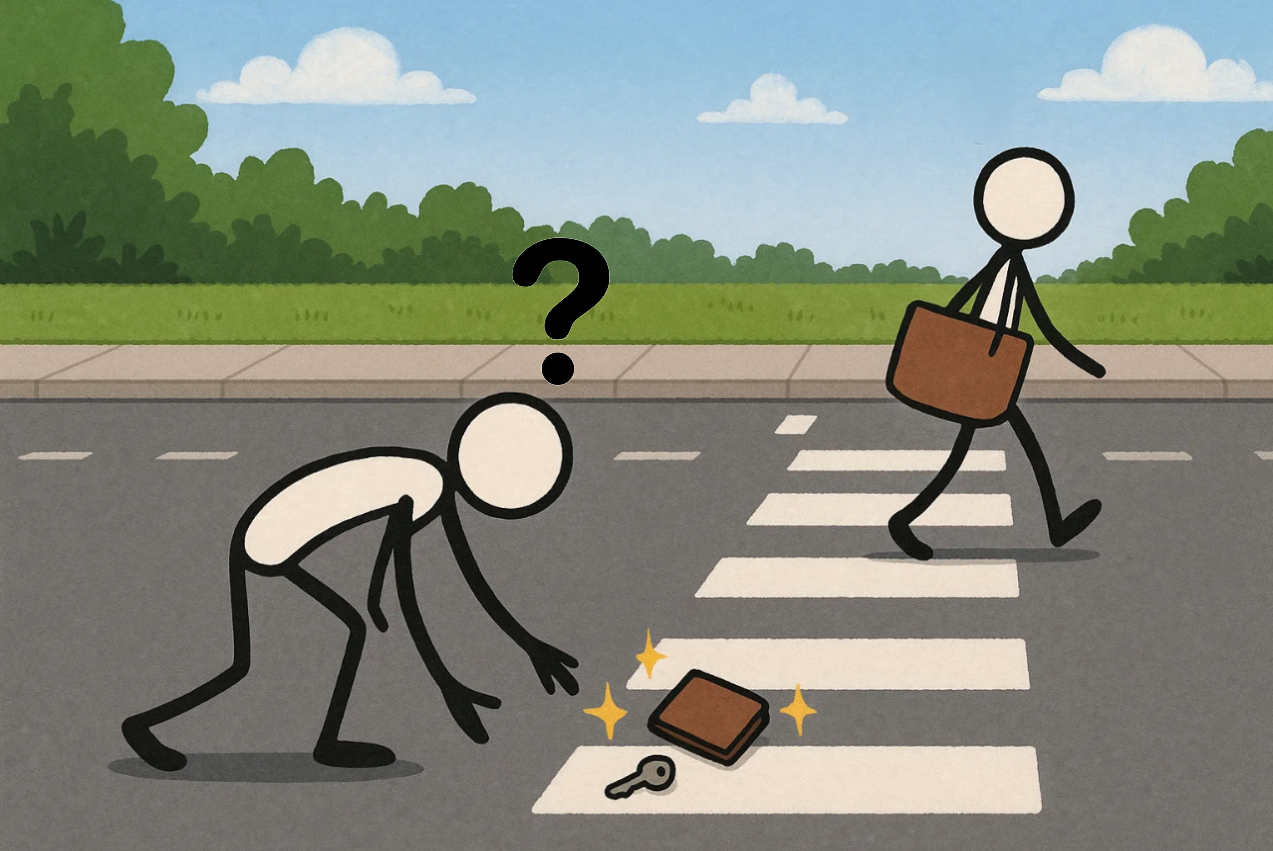}
    \caption{An individual discovers a wallet near its owner, confronting the moral dilemma of returning it or keeping it.}
    \label{fig:stim2}
\end{subfigure}
\hfill
\begin{subfigure}{0.32\textwidth}
    \includegraphics[width=\linewidth]{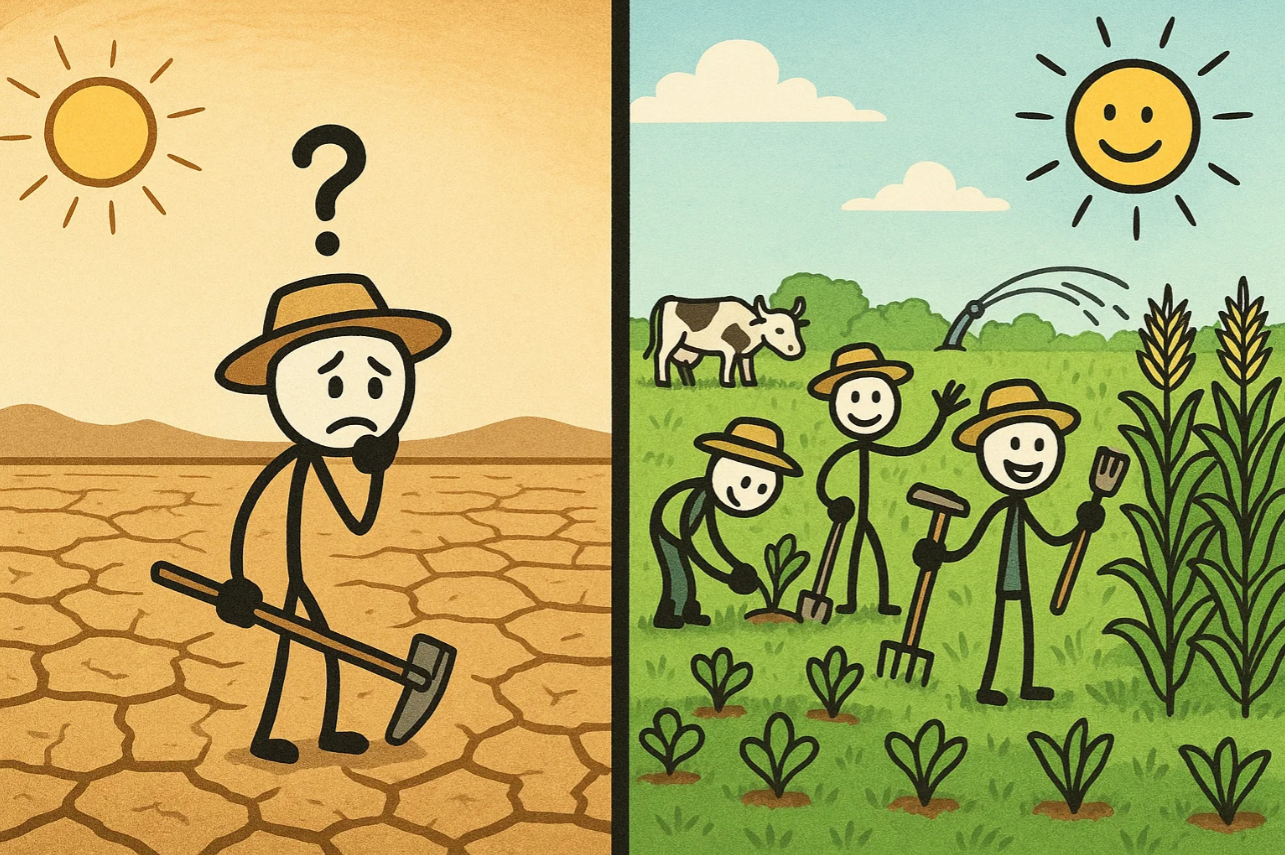}
    \caption{Farmers face the ethical dilemma of allocating limited resources to sustain cattle during a severe drought.}
    \label{fig:stim3}
\end{subfigure}
\caption{Exemplar pictorial moral dilemmas used as stimuli. Full set of 13 dilemmas available upon request.}
\label{fig:stimuli_examples}
\end{figure}

\section{Attribute B: Confidence Factor Modeling}

Attribute B in our Z-number framework quantifies the confidence participants associate with their moral decision-making. After each response, participants self-reported their confidence on a continuous sliding scale $r_{i,j} \in [1,10]$, indicating how sure they were that their chosen action was wise.

\subsection{Kernel Density Estimation and Confidence Segmentation}

To transform the continuous confidence scores $B_i \in [0,1]$ into interpretable linguistic categories, we employed a nonparametric Gaussian Kernel Density Estimation (KDE) approach. This allowed for data-driven segmentation without enforcing arbitrary binning thresholds.

\begin{figure}[H]
\centering
\includegraphics[width=\linewidth]{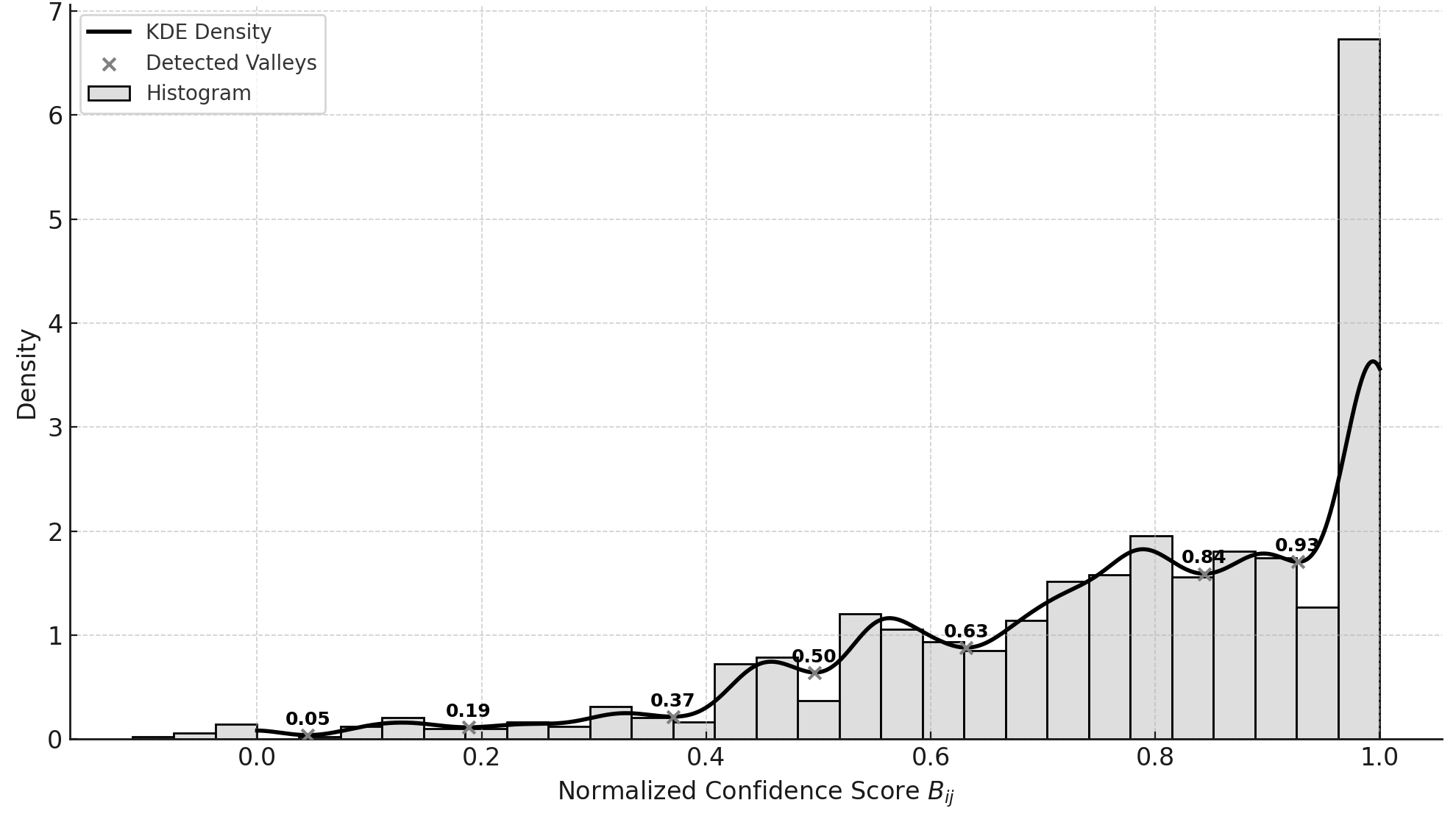} 
\caption{Gaussian KDE of all normalized confidence ratings $B_{i,j}$ with automatically detected valley points (local minima). These serve as fuzzy segmentation boundaries for Attribute B labels.}
\label{fig:kde attribute B.png}
\end{figure}

Given $N \times J$ normalized values $B_{i,j}$ across all participants and stimuli, the KDE estimator $\hat{f}_B(x)$ of the underlying confidence distribution is defined in \eqref{kde}:
\begin{equation}\label{kde}
\hat{f}_B(x) = \frac{1}{Nh} \sum_{i=1}^N \sum_{j=1}^J K\left( \frac{x - B_{i,j}}{h} \right),
\end{equation}
where $K(\cdot)$ is a Gaussian kernel as shown in \eqref{gau}:
\begin{equation} \label{gau}   
K(u) = \frac{1}{\sqrt{2\pi}} e^{- \frac{1}{2} u^2},
\end{equation}
and $h$ is the bandwidth controlling the smoothness of the estimated density. We set $h = 0.12$ based on Silverman’s rule of thumb and empirical tuning.

The resulting $\hat{f}_B(x)$ provides a continuous, differentiable estimate of the empirical density of confidence levels. To identify natural breakpoints in the population’s confidence patterns, we detect local minima (valleys as shown in \eqref{valley}) in the negative KDE:
\begin{equation}\label{valley}  
\text{Valleys} = \arg\max_x \left[ -\hat{f}_B(x) \right].
\end{equation}
This method yielded seven distinct valleys:
\[
0 = t_0 < t_1 < t_2 < t_3 < t_4 < t_5 < t_6 < t_7 < t_8 = 1,
\]
thus partitioning the interval $[0,1]$ into seven contiguous regions.

Each region was mapped to a linguistic descriptor based on modal strength commonly used in psychometric confidence expressions:
\[
\begin{aligned}
\{ & \text{Perhaps}, \text{Possibly}, \text{Probably}, \text{Supposedly}, \\
   & \text{Expectedly}, \text{Decisively}, \text{Certainly} \}.
\end{aligned}
\]

This 7-tier confidence taxonomy enables finer granularity in the fuzzy interpretation of self-assessed moral certainty. The KDE plot with labeled valleys is illustrated in supplementary.

\subsection{Fuzzy Membership Assignment}
Each modeled using either a triangular or trapezoidal membership function depending on position in the confidence spectrum. The outermost sets, \textit{Perhaps} and \textit{Certainly}, are modeled as trapezoidal functions, while all intermediate sets are modeled as symmetric triangular functions. Let the extracted threshold points be $t_0 = 0$, $t_1$, $\dots$, $t_7$, and $t_8 = 1$.

Let the threshold values be:
\[
\begin{aligned}
t_0 &= 0, \quad t_1 = 0.05, \quad t_2 = 0.19, \quad t_3 = 0.37, \\
t_4 &= 0.50, \quad t_5 = 0.63, \quad t_6 = 0.84, \quad t_7 = 0.93, \quad t_8 = 1.0
\end{aligned}
\]

The piecewise functions are defined as follows:

\begin{equation}
\mu_{\text{Perhaps}}(x) =
\begin{cases}
1, & x \leq t_1 \\
\frac{t_2 - x}{t_2 - t_1}, & t_1 < x < t_2 \\
0, & x \geq t_2
\end{cases}
\end{equation}

\begin{equation}\label{poss}
\mu_{\text{Possibly}}(x) =
\begin{cases}
0, & x \leq t_1 \\
\frac{x - t_1}{t_2 - t_1}, & t_1 < x \leq t_2 \\
\frac{t_3 - x}{t_3 - t_2}, & t_2 < x < t_3 \\
0, & x \geq t_3
\end{cases}
\end{equation}

\begin{equation}\label{prob}
\mu_{\text{Probably}}(x) =
\begin{cases}
0, & x \leq t_2 \\
\frac{x - t_2}{t_3 - t_2}, & t_2 < x \leq t_3 \\
\frac{t_4 - x}{t_4 - t_3}, & t_3 < x < t_4 \\
0, & x \geq t_4
\end{cases}
\end{equation}

\begin{equation}\label{supp}
\mu_{\text{Supposedly}}(x) =
\begin{cases}
0, & x \leq t_3 \\
\frac{x - t_3}{t_4 - t_3}, & t_3 < x \leq t_4 \\
\frac{t_5 - x}{t_5 - t_4}, & t_4 < x < t_5 \\
0, & x \geq t_5
\end{cases}
\end{equation}

\begin{equation}\label{exp}
\mu_{\text{Expectedly}}(x) =
\begin{cases}
0, & x \leq t_4 \\
\frac{x - t_4}{t_5 - t_4}, & t_4 < x \leq t_5 \\
\frac{t_6 - x}{t_6 - t_5}, & t_5 < x < t_6 \\
0, & x \geq t_6
\end{cases}
\end{equation}

\begin{equation}\label{dec}
\mu_{\text{Decisively}}(x) =
\begin{cases}
0, & x \leq t_5 \\
\frac{x - t_5}{t_6 - t_5}, & t_5 < x \leq t_6 \\
\frac{t_7 - x}{t_7 - t_6}, & t_6 < x < t_7 \\
0, & x \geq t_7
\end{cases}
\end{equation}

\begin{equation}\label{cer}
\mu_{\text{Certainly}}(x) =
\begin{cases}
0, & x \leq t_6 \\
\frac{x - t_6}{t_7 - t_6}, & t_6 < x \leq t_7 \\
1, & x > t_7
\end{cases}
\end{equation}

\begin{figure}[!t]
    \centering
    \includegraphics[width=0.8\textwidth]{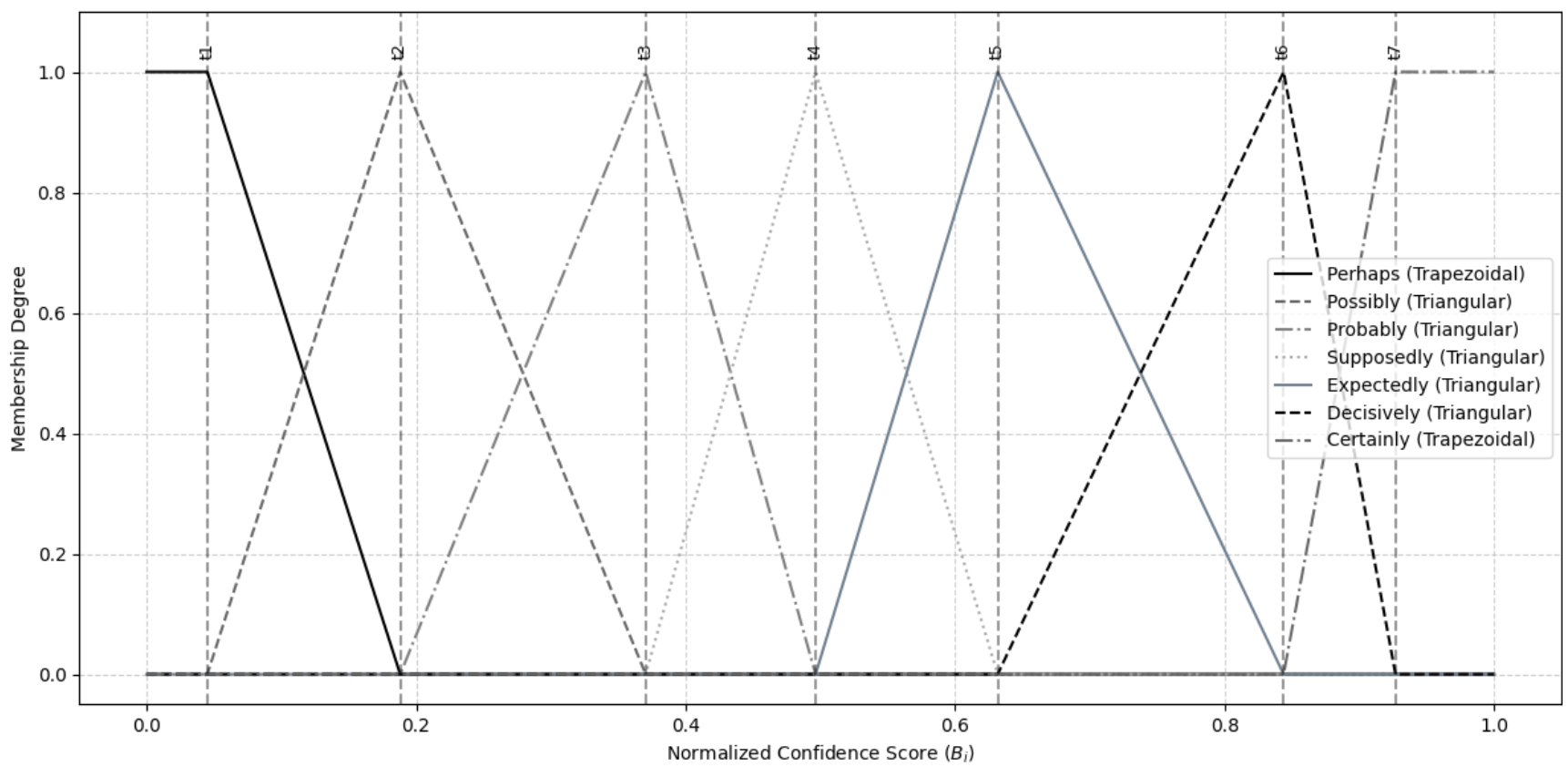} % Replace with your actual MF plot filename
    \caption{Fuzzy membership functions $\mu_\ell(x)$ over normalized confidence scores $x \in [0,1]$. Trapezoidal shapes are used for edge labels (e.g., \textit{Certainly}), and triangular shapes for intermediate categories. Thresholds $t_1$ through $t_7$ were extracted from KDE valley points.}
    \label{fig:attribute B membership function.png}
\end{figure}

Figure~\ref{fig:attribute B membership function.png} shows the full suite of membership functions spanning the entire $[0,1]$ range.

\subsection{Algorithmic Procedure for Attribute B Inference}

Algorithm~\ref{alg:attributeB} summarizes the full computation pipeline, from raw confidence ratings to final Attribute B labels.

\begin{algorithm}[H]
\small
\caption{Attribute B Computation (Confidence Factor)}
\label{alg:attributeB}
\begin{algorithmic}[1]
\STATE \textbf{Input:} Raw ratings $ r_{i,j} \in [1,10]$, 
       with $N = 100$ participants, $J = 13$ dilemmas
\FOR{$i = 1$ to $N$}
    \FOR{$j = 1$ to $J$}
        \STATE Normalize: $ B_{i,j} \gets (r_{i,j} -1)/9$
    \ENDFOR
    \STATE Compute participant mean: 
          $ B_i \gets \frac{1}{J} \sum_{j=1}^J B_{i,j}$
\ENDFOR
\STATE Apply KDE over $\{ B_i \}$:
       $ f_B(x) \gets \frac{1}{Nh} \sum_{i=1}^N K\left( \frac{x - B_i}{h} \right)$
\STATE Detect valleys (local minima) in $f_B(x)$ using peak detection on $-f_B(x)$,
       yielding natural boundaries $t_1, t_2, \dots, t_M$ over $x \in [0,1]$
\STATE Define fuzzy membership functions 
       $\mu_\ell(B)$, where 
       $\ell \in \{\text{Perhaps}, \text{Possibly}, \text{Probably},
       \text{Supposedly},$
\STATE \hspace{0.5cm}$\text{Expectedly}, \text{Decisively}, \text{Certainly}\}$ 
       using valley-derived $t_M$ boundaries
\FOR{$i = 1$ to $N$}
    \STATE Assign label:
          $ \ell_i^* \gets \arg\max_\ell \mu_\ell(B_i)$
\ENDFOR
\end{algorithmic}
\end{algorithm}

The procedure above systematically converts participant-level confidence scores into interpretable fuzzy labels based on their position within the empirical distribution. 

\section{Fuzzy Inference Rules}

\begin{table}[H]
\centering
\caption{Summary of Fuzzy Inference Rules (R1--R21) for Wisdom-Level Attribute A}
\label{tab:fuzzy_inference_rules_21}
\footnotesize
\renewcommand{\arraystretch}{1.3}
\setlength{\tabcolsep}{4pt}
\begin{tabular}{|c|p{6.4cm}|c|p{5.8cm}|}
\hline
\textbf{Rule ID} & \textbf{IF Antecedents (Fuzzy Conditions)} & \textbf{THEN Wisdom Category} & \textbf{Empirical Justification} \\
\hline
R1 & $\geq 3$ of $\{\text{IH}_{high}, \text{PT}_{high}, \text{Ref}_{high}, \text{Pro}_{high}, \text{REA}_{high}\} \geq 0.5$ \newline AND $(\text{Pro}_{high} \lor \text{REA}_{high}) \geq 0.5$ & High Wisdom & Moral grounding, through concern or action, anchors wise reasoning (Ardelt, 2003; Grossmann et al., 2012). \\
\hline
R2 & $\text{IH}_{high} \land \text{PT}_{high} \land \text{Ref}_{high}$ & High Wisdom & Metacognitive humility, perspective-taking, and deep reflection jointly predict maximal wisdom (Grossmann et al., 2020; Brienza \& Grossmann, 2017). \\
\hline
R3 & $\text{PT}_{high} \land \text{Pro}_{high} \land \text{REA}_{high}$ & High Wisdom & Social-cognitive skill plus moral concern and action foster collective well-being and wisdom (Pinto et al., 2024; Glück et al., 2019). \\
\hline
R4 & $\text{IH}_{high} \land (\text{Pro}_{high} \lor \text{REA}_{high})$ & High Wisdom & Strong humility combined with moral grounding suffices for wise judgment (Glück et al., 2019). \\
\hline
R5 & $\text{PT}_{low} \land \text{IH}_{high} \land (\text{Pro}_{high} \lor \text{REA}_{high} \lor \text{Ref}_{high})$ & High Wisdom & Cultural contexts can allow adaptive wisdom despite low perspective-taking (Brienza et al., 2018). \\
\hline
R6 & $\text{REA}_{low} \land \text{IH}_{high} \land \text{PT}_{high} \land \text{Pro}_{high}$ & High Wisdom & Low action frequency can be offset by strong moral intent and perspective skills (Pinto et al., 2024). \\
\hline
R7 & $\geq 3$ of $\{\text{IH}_{mod}, \text{PT}_{mod}, \text{Ref}_{mod}, \text{Pro}_{mod}, \text{REA}_{mod}\} \geq 0.5$ \newline AND no $\mu_{\cdot,low} \geq 0.5$ & Moderate Wisdom & Non-exceptional but balanced profiles reflect intermediate wise reasoning (Webster, 2003; Glück et al., 2019). \\
\hline
R8 & $\text{IH}_{mod} \land \text{PT}_{mod} \land \text{Ref}_{mod}$ & Moderate Wisdom & Moderate humility, perspective-taking, and reflectiveness yield practical wisdom (Baltes \& Staudinger, 2000; Sternberg, 1998). \\
\hline
R9 & $\text{Pro}_{mod} \land \text{REA}_{mod} \land \text{Ref}_{mod}$ & Moderate Wisdom & Balanced moral concern and reflective action support pragmatic wisdom (Ardelt, 2003; Glück et al., 2019). \\
\hline
R10 & $(\text{IH}_{high} \lor \text{PT}_{high}) \land$ exactly one of $\{\text{Pro},\text{REA},\text{Ref}\} = low$ & Moderate Wisdom & A single deficit can lower wisdom from high to moderate; compensatory balance is key (Sternberg, 1998; Brienza et al., 2018). \\
\hline
R11 & $\text{PT}_{mod} \land \text{Pro}_{mod} \land \text{REA}_{low} \land \text{Ref}_{high}$ & Moderate Wisdom & Strong reflection without consistent action produces sound but limited wisdom (Glück et al., 2019; Ardelt, 2003). \\
\hline
R12 & $\text{IH}_{mod} \land \text{PT}_{high} \land (\text{Pro}_{mod} \lor \text{REA}_{mod})$ & Moderate Wisdom & High perspective-taking with moderate moral concern/action yields solid, not maximal, wisdom (Grossmann, 2017). \\
\hline
R13 & $\text{IH}_{low} \land \text{PT}_{low} \land \text{Ref}_{low}$ & Low Wisdom & Low humility, poor perspective-taking, and weak reflection strongly predict unwise decisions (Walsh, 2015; Grossmann et al., 2020). \\
\hline
R14 & $\text{Pro}_{low} \land \text{REA}_{low} \land \text{Ref}_{low}$ & Low Wisdom & Minimal moral concern, weak empathetic action, and low reflectiveness undermine wisdom (Pinto et al., 2024). \\
\hline
R15 & $\text{Pro}_{low} \land \text{REA}_{low}$ (irrespective of others) & Low Wisdom & Lack of moral grounding and compassionate action constrains wisdom (Ardelt, 2003; Grossmann, 2017). \\
\hline
R16 & $\text{PT}_{low} \land \text{IH}_{low}$ & Low Wisdom & Weak perspective-taking and low humility correlate with egocentric reasoning (Brienza et al., 2018). \\
\hline
R17 & all five components = low & Low Wisdom & Broad, consistent deficits lead to persistently poor wise reasoning (Grossmann et al., 2020). \\
\hline
R18 & all five components = moderate & Moderate Wisdom & Balanced moderate levels across attributes align with pragmatic, context-sensitive reasoning (Sternberg, 1998). \\
\hline
R19 & all five components = high & High Wisdom & Consistent high levels across wisdom attributes indicate maximally wise reasoning (Grossmann et al., 2020). \\
\hline
R20 & $\text{IH}_{low} \land \text{Pro}_{low}$ & Low Wisdom & Low humility with low moral concern predicts self-serving, unwise reasoning (Ardelt, 2003). \\
\hline
R21 & Default: no rule fires with strength $\geq 0.5$ & Moderate Wisdom & Conservative fallback maintains interpretability in ambiguous profiles (Sternberg, 1998). \\
\hline
\end{tabular}
\end{table}

%===================================================

%=====================================================================